\pdfoutput=1

\documentclass[11pt]{article}

\usepackage[final]{acl}

\usepackage{times}
\usepackage{latexsym}

\usepackage{amsmath,amsfonts,bm}

\def\eqref#1{equation~\ref{#1}}

\def\1{\bm{1}}

\DeclareMathAlphabet{\mathsfit}{\encodingdefault}{\sfdefault}{m}{sl}
\SetMathAlphabet{\mathsfit}{bold}{\encodingdefault}{\sfdefault}{bx}{n}

\usepackage{booktabs}
\usepackage{multirow}
\usepackage{multicol}
\usepackage{graphicx}
\usepackage{pbox}
\usepackage{float}

\usepackage[ruled, vlined]{algorithm2e}
\usepackage{algorithmic}

\usepackage[T1]{fontenc}

\usepackage[utf8]{inputenc}

\usepackage{microtype}

\usepackage{inconsolata}

\usepackage{graphicx}

\title{Neural Topic Modeling with Large Language Models in the Loop}

\author{
 \textbf{Xiaohao Yang\textsuperscript{1}},
 \textbf{He Zhao\textsuperscript{2}\Thanks{Corresponding authors: Lan Du, He Zhao}},
 \textbf{Weijie Xu\textsuperscript{3}},
 \\
 \textbf{Yuanyuan Qi\textsuperscript{1}},
 \textbf{Jueqing Lu\textsuperscript{1}},
 \textbf{Dinh Phung\textsuperscript{1}},
 \textbf{Lan Du\textsuperscript{1}$^{\ast}$}
\\
\\
 \textsuperscript{1}Faculty of IT, Monash University, Australia \\
 \textsuperscript{2}CSIRO's Data61, Australia \
 \textsuperscript{3}Amazon, America 
\\
\texttt{\{xiaohao.yang,yuanyuan.qi,jueqing.lu,dinh.phung,lan.du\}@monash.edu}\\
\texttt{he.zhao@data61.csiro.au, \ weijie.xu@amazon.com}
}

\begin{document}
\maketitle
\begin{abstract}
Topic modeling is a fundamental task in natural language processing, allowing the discovery of latent thematic structures in text corpora. While Large Language Models (LLMs) have demonstrated promising capabilities in topic discovery, their direct application to topic modeling suffers from issues such as incomplete topic coverage, misalignment of topics, and inefficiency. To address these limitations, we propose LLM-ITL, a novel LLM-in-the-loop framework that integrates LLMs with Neural Topic Models (NTMs). In LLM-ITL, global topics and document representations are learned through the NTM. Meanwhile, an LLM refines these topics using an Optimal Transport (OT)-based alignment objective, where the refinement is dynamically adjusted based on the LLM's confidence in suggesting topical words for each set of input words. With the flexibility of being integrated into many existing NTMs,
the proposed approach enhances the interpretability of topics while preserving the efficiency of NTMs in learning topics and document representations. Extensive experiments demonstrate that LLM-ITL helps NTMs significantly improve their topic interpretability while maintaining the quality of document representation. Our code and datasets are available at \href{https://github.com/Xiaohao-Yang/LLM-ITL}{https://github.com/Xiaohao-Yang/LLM-ITL}
\end{abstract}

\section{Introduction}
Topic modeling is an essential task in natural language processing that uncovers hidden thematic structures within text collections in an unsupervised way. The ability to automatically extract topics has proven to be invaluable across a range of disciplines, such as bioinformatics \citep{Liu2016}, marketing research \citep{reisenbichler2019topic}, and information retrieval \citep{10.1007/978-3-642-00958-7_6}. 
Topic models are conventionally based on probabilistic frameworks such as Latent Dirichlet Allocation (LDA) \citep{10.5555/944919.944937} and its hierarchical Bayesian extensions~\citep{paisley2015nested,gan2015learning,zhou2016augmentable,zhao2018inter,zhao2018dirichlet}, which generate a set of interpretable global topics, each represented as a distribution over vocabulary terms. These topics are then used to represent individual documents as mixtures of topics, providing a structured and interpretable view of the corpus. Recently, research on Neural Topic Models (NTMs) \citep{ijcai2021p638,10.1145/3507900,ABDELRAZEK2023102131,wu2024survey} has been popular, which uses deep neural networks to model document-topic distributions, enabling more expressive and flexible representations compared to their probabilistic counterparts.

While Large Language Models (LLMs)~\citep{openai2023chatgpt,touvron2023llama,touvron2023llama2} have redefined the landscape of natural language processing, topic models can still be valuable tools for text analysis. 
Specifically, LLMs can provide a fine-grained understanding of a document; however, given a large collection of domain-specific documents, 
topic models are more suitable to obtain a clear global view of the topics in a more interpretable way with much less computational cost.
Unsurprisingly, it has been a trending research direction to use LLMs to improve topic modeling \citep{6fe80f19eed94ff2b92c510b6e2670e8,10386113,pham-etal-2024-topicgpt,mu-etal-2024-large,doi-etal-2024-topic,chang2024enhanced}.
Despite the promising performance of these initial studies, most existing methods involve prompting LLMs to generate topics for each document in the corpus, which may lead to several limitations.
As LLMs are asked to focus on a document individually, they may be unable to cover global topics across all the documents in the corpus \citep{doi-etal-2024-topic}, which is critical in topic modeling. 
Moreover, although LLMs excel at capturing local context, they usually struggle with long documents with multiple interrelated topics, which may evolve or shift throughout the text. With their limited window of focus, LLMs may miss key topics of a document that are necessary to fully understand its content.
Finally, it is computationally expensive as LLMs have to do inference for every document in the corpus or subsets; thus, existing methods usually scale poorly with large datasets.

To overcome the aforementioned limitations, we propose \textbf{LLM-ITL}, a framework that integrates LLMs into NTMs and enhances the overall quality and interpretability of the learned topics, while maintaining computational efficiency. 
Specifically, to enhance the interpretability of the topics learned by an NTM, we introduce an LLM-based refinement step. The representative words for each topic, as generated by the NTM, are provided to the LLM, which suggests improved words that better describe the semantic meaning of the topic. The refinement process is guided by a novel plug-in objective based on Optimal Transport (OT), which ensures that the topics learned by the NTM align closely with the LLM’s refinements.
Additionally, to mitigate potential hallucinations from the LLM (i.e., the generation of inaccurate or irrelevant suggestions) for hard or incoherent topics, we introduce a confidence-weighted mechanism that adjusts the influence of the LLM’s suggestions based on their confidence scores. Finally, to balance learning from the input corpus by the NTM and refinement by the LLM, a warm-up phase is adopted to ensure that the NTM component learns corpus-relevant topics and document representations before the LLM's refinement is applied.
Our proposed LLM-ITL framework offers the following key contributions:
\begin{itemize}
    \item \textbf{Boosting topic quality while maintaining document representation quality}: With the LLM’s refined topics and OT-based alignment, the topics generated are more interpretable and semantically coherent. At the same time, LLM-ITL ensures that the document-topic distributions, as learned by the NTM, remain high-quality and reflective of the document’s content.
    \item \textbf{Efficiency in using LLMs}: Unlike most existing LLM-based approaches that rely on document-level LLM analysis, LLM-ITL uses LLMs at the word level, significantly reducing computational overhead. 
    \item \textbf{Flexibility}: LLM-ITL is a modular framework that can integrate with a variety of NTMs and LLMs, offering flexibility in model selection depending on the application and computational constraints.
\end{itemize}

\begin{figure*}[!ht]
    \centering
    \includegraphics[width=0.9\textwidth]{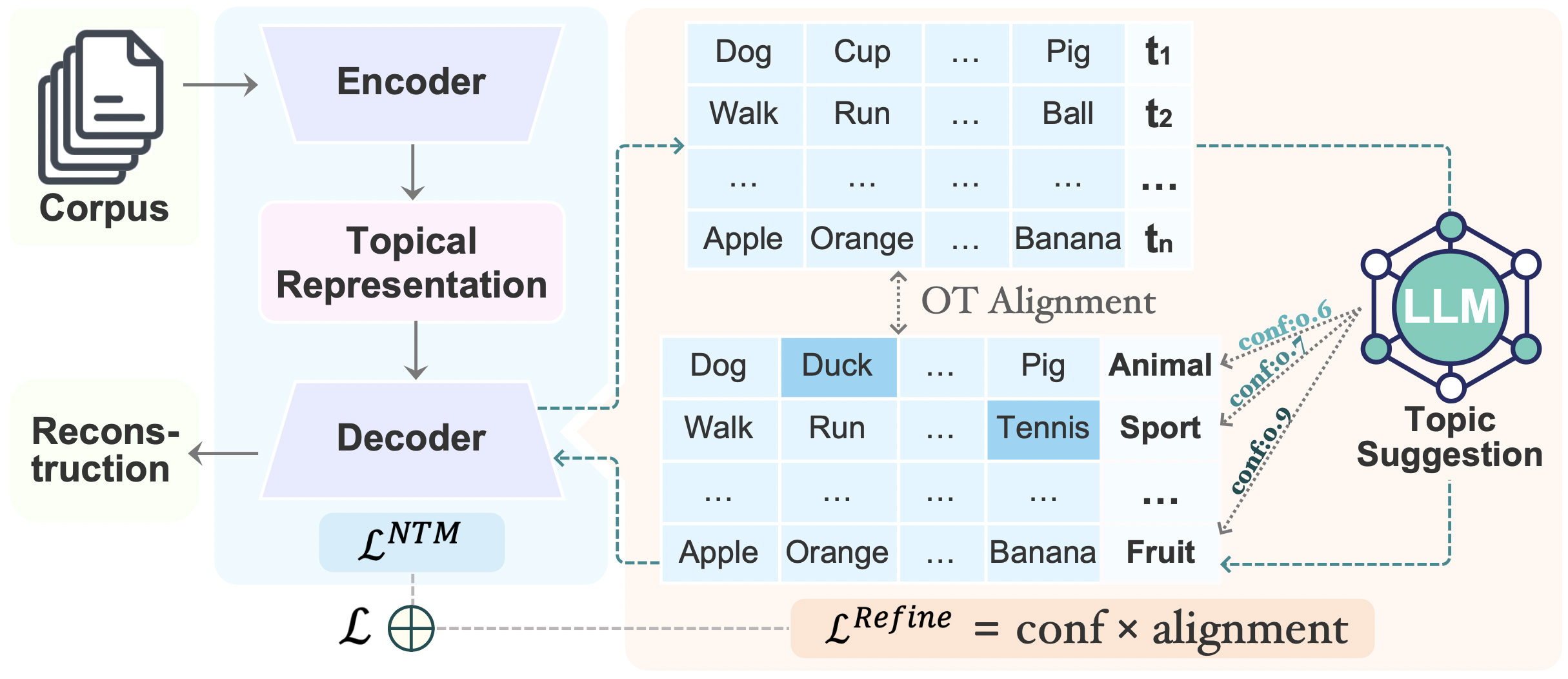}
  \caption{LLM-ITL overview. The topics and document representations are learned by the Neural Topic Model (NTM) component. After a warm-up stage, a Large Language Model (LLM) suggests better topic words for the learned topics (e.g., decoder) from the NTM. An Optimal Transport (OT)-based topic alignment objective is proposed to align the word distribution between the topics from the NTM and those suggested by the LLM. The alignment is further weighted by the confidence of the LLM in providing the suggestions. This confidence-weighted topic refinement objective is plugged into the standard training of an NTM as the overall objective of LLM-ITL.}\label{fig:flow}
\end{figure*}

\section{Background}
\subsection{Problem Setup for Topic Modeling}
Given a document collection $\mathcal{D}:=\{\bm{d}_1,\ldots,\bm{d}_N\}$, a topic model learns to discover a set of global topics $\mathcal{T}:=\{\bm{t}_1,\ldots,\bm{t}_K\}$, each of which  is a distribution over the $V$ vocabulary words $\bm{t} \in \Delta^{V}$ ($\Delta$ denotes the probability simplex). Ideally, each topic represents a semantic concept that can be interpreted with its top-weighted words. At the document level, the topic model represents each document as a distribution over the $K$ topics, i.e., $\bm{z}\in \Delta^K$, which indicates the topic proportion of each topic within the document. The interpretability of topic models derives from both the corpus-level topics $\mathcal{T}$, and the document-level topical representation $\bm{z}$ for each document.

\subsection{Neural Topic Models}
A Neural Topic Model (NTM) is typically trained by modeling $p(\bm{z}|\bm{x})$ and $p(\bm{x}|\bm{z})$, where $\bm{x} \in \mathbb{N}^{V}$ represents the Bag-of-Words (BoWs) of a document. NTMs, which employ deep neural networks for topic modeling, are commonly based on Variational Auto-Encoders (VAEs) \citep{Kingma2014} and Amortized Variational Inference (AVI) \citep{10.5555/3044805.3045035}. For VAE-NTMs, $p(\bm{x}|\bm{z})$ is modeled by a decoder network $\bm{\phi}$, i.e., $\bm{x} := \bm{\phi}(\bm{z})$. The posterior $p(\bm{z}|\bm{x})$ is approximated by $q(\bm{z}|\bm{x})$, which is modeled by an encoder network $\bm{\theta}$, i.e., $\bm{z} := \bm{\theta}(\bm{x})$.
The training objective of VAE-NTMs is to maximize the Evidence Lower Bound (ELBO):

\vspace{-0.9em}
\small
\begin{equation} \label{ntm_loss}
\resizebox{0.48\textwidth}{!}{$
\max\limits_{\bm{\theta},\bm{\phi}}{(\mathbb{E}_{q_{\bm{\theta}}(\bm{z}|\bm{x})}[\log{p_{\bm{\phi}}(\bm{x}|\bm{z})}]}-\mathbb{KL}[q_{\bm{\theta}}(\bm{z}|\bm{x})\parallel p(\bm{z})]),
$}
\end{equation}
\normalsize
where the first term encourages the reconstruction of the document, and the second is the Kullback–Leibler divergence between the approximate posterior and the prior distribution. By implementing a single linear layer for the decoder $\bm{\phi}\in \mathbb{R}^{V\times K}$, the $k$-th topic distribution $\bm{t}_k$ can be obtained by normalizing the $k$-th column of the decoder's weight matrix:

\vspace{-1em}
\small
\begin{equation}\label{eq_t}
    \bm{t}_{k}:=\text{softmax}(\bm{\phi}_{:,k})^\text{T}.
\end{equation}
\normalsize
Its topic words $\bm{w}_k$ are obtained by taking the top-weight words of $\bm{t}_k$, written as:

\vspace{-0.3em}
\small
\begin{equation}\label{eq_w}
    \bm{w}_k := \mathcal{V}[f_{\text{topn}}(\bm{t}_k, N)],
\end{equation}
\normalsize
where $f_{\text{topn}}(\bm{a},N)$ defines a function that returns the indices of the top-$N$ values of vector $\bm{a}$; $\mathcal{V}$ denotes the vocabulary set of the corpus.

\subsection{Optimal Transport}
Optimal Transport (OT) has been widely used for comparing probability distributions \citep{10.5555/2999792.2999868,10.5555/2969442.2969469,seguy2018large,peyre2019computational}, which extensive applications in machine learning and related areas~\cite{ge2021ota,nguyen2021most,wanrepresenting2022,danlearning2022,buiunified,vuong2023vector,zhao2023transformed,ye2024ptarl,vo2024optimal,gao2024distribution,ren2024modality}. Let $\mu(\bm{x}, \bm{a}) := \sum_{i=1}^N a_i \delta_{x_i}$ and $\mu(\bm{y}, \bm{b}) := \sum_{j=1}^{M} b_j \delta_{y_j}$ be two discrete distributions, where $\bm{a} := [a_1, \ldots, a_N]$ and $\bm{b} := [b_1, \ldots, b_M]$ are the probability vectors; $\bm{x} := \{x_1, \ldots, x_N\}$ and $\bm{y} := \{y_1, \ldots, y_M\}$ are the supports of these two distributions. The OT distance between $\mu(\bm{x}, \bm{a})$ and $\mu(\bm{y},\bm{b})$ is obtained by finding the optimal transport plan $\bm{P}^*$ that transports the probability mass from $\bm{a}\in \Delta^{N}$ to $\bm{b}\in\Delta^{M}$, written as following:

\vspace{-1em}
\small
\begin{equation}\label{ot}
    d_{\text{OT}}(\mu(\bm{x}, \bm{a}), \mu(\bm{y}, \bm{b}))  := \min_{\bm{P}} \sum_{i=1}^{N} \sum_{j=1}^{M} C_{i,j} P_{i, j}, 
\end{equation}
\normalsize
\vspace{-1.0em}

\noindent subject to $\sum_{j=1}^{M} P_{i,j}  = a_i, \ \forall i=1,\ldots,N$ and  
    $\sum_{i=1}^{N} P_{i,j}  = b_{j}, \ \forall j=1,\ldots,M$.
Here, $\bm{P}\in\mathbb{R}_{\ge 0}^{N\times M}$ is the transport plan, with entry $P_{i,j}$ indicating the amount of probability mass moving from $a_i$ to $b_j$; $\bm{C}\in\mathbb{R}_{\ge 0}^{N\times M}$ denotes the cost matrix, with entry $C_{i,j}$ specifying the distance between supports $x_i$ and $y_j$. Various OT solvers \citep{10.5555/3546258.3546336} have been proposed to compute the OT distance.

\section{Method}
In this work, we propose LLM-ITL, an LLM-in-the-loop framework that efficiently integrates the LLM with the training of NTMs, offering a more interpretable and comprehensive topic modeling pipeline. An overview of LLM-ITL is illustrated in Figure \ref{fig:flow}. LLM-ITL involves the following key components: LLM-based topic suggestion, OT distance for topic alignment, and confidence-weighted topic refinement.

\begin{figure*}[t]
    \centering
    \includegraphics[width=0.95\textwidth]{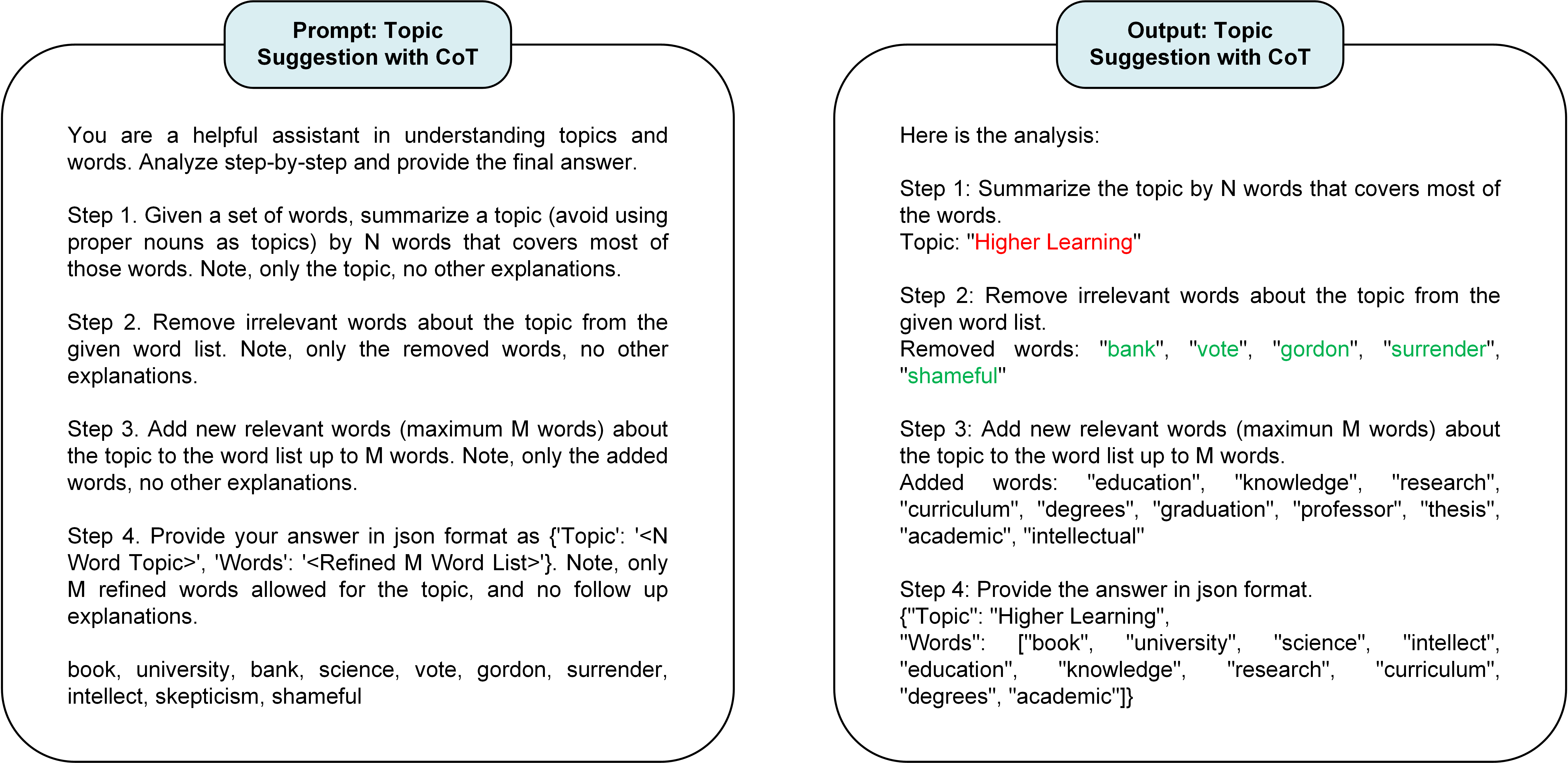}
  \caption{Prompt and output of topic suggestion with CoT. We take the product of token probabilities of topic label (e.g., words in \textbf{red} color) as the Label Token Probability. We take the proportion of intruders (e.g., words in \textbf{green} color) as the Word Intrusion Confidence. $N=2$ and $M=10$ in this example.}
  \label{fig:prompt_refine}
\end{figure*}

\subsection{LLM-based Topic Suggestion}
During the training of an NTM, 
it typically generates a set of topics, where each topic is represented by a distribution over words, with the highest-probability words forming the core ``meaning'' of the topic. While these words offer a rough semantic grouping, they often lack clarity or precision, leading to difficulties in interpretation. For instance, topics may contain words that are too general, too specific, or semantically ambiguous, making it hard for users to derive clear labels or understand the thematic focus of the topic.

To address this, LLM-ITL proposes to use LLMs to suggest better words or labels that more clearly express the same underlying concept. The LLM is prompted with the top words from each topic, and it generates two outputs: a topic label, which is a concise and interpretable summary of the topic; and a set of refined topic words, which better represent the underlying semantic concept of the topic. This process capitalizes on the LLM's ability to grasp language nuances and provide more semantically rich suggestions for the topic. The LLM’s extensive pre-training on diverse and large datasets allows it to capture subtle relationships between words that may not be apparent in the purely statistical or neural-based methods employed by NTMs.

To obtain the topic label and refined words in a structured manner, \textit{chain-of-thought (CoT) prompting}~\citep{10.5555/3600270.3602070} is employed. CoT prompting encourages the LLM to reason step-by-step through the task, ensuring that it carefully considers the topic words before generating a label and refinement. The LLM’s output sequence $\bm{s}$ includes both the \textit{topic label} and \textit{refined words}, extracted as follows for each set of topic words:

\vspace{-1.0em}
\small
\begin{align}\label{llm_opt}
\bm{s} & := \bm{\theta}^{\text{llm}}(\text{Prompt}(\bm{w})), \notag \\
\text{Topic label } \bm{w}^l &: (s_{\text{start of label}}, \ldots, s_{\text{end of label}}), \notag\\
\text{Refined words } \bm{w}' &: (s_{\text{start of words}}, \ldots, s_{\text{end of words}}),
\end{align}
\normalsize
\vspace{-1.2em}

\noindent where $\bm{w}$ represents the original topic words; $\bm{\theta}^{\text{llm}}$ denotes the LLM model; the topic label $\bm{w}^l$ and the refined words $\bm{w}'$ are extracted as subsequences from the LLM's output $\bm{s}$. Notably, the words in $\bm{w}'$ are further filtered based on the corpus vocabulary to ensure that no out-of-vocabulary (OOV) words are included. Our used prompt is illustrated in Figure \ref{fig:prompt_refine}. A study of prompt variants is provided in Appendix \ref{appendix:prompt_variants}.

\subsection{OT-based Topic Alignment}
A key innovation in LLM-ITL is the use of Optimal Transport (OT) distance to align the topic word distributions generated by the NTM with the refined topic word distributions provided by the LLM. OT is a mathematical framework that computes the ``cost'' of transforming one probability distribution into another, and has shown its effectiveness in measuring the alignment between two sets of words~\citep{pmlr-v37-kusnerb15,10.1162/tacl_a_00744}.

Formally, given a set of original topic words $\bm{w} := \{w_1, w_2, \ldots, w_N\}$ with probability vector $\bm{t} := [t_1, t_2, \ldots, t_N]$ as obtained by Eq. \ref{eq_w} and Eq. \ref{eq_t}, respectively; as well as refined topic words $\bm{w}' := \{w'_1, w'_2, \ldots, w'_M\}$ with probability vector $\bm{u} := [u_1, u_2, \ldots, u_M]$\footnote{We assume each of the refined topic words from the LLM is equally important, thus $\bm{u}$ is a uniform probability vector.} from the LLM, the OT distance between these two word distributions can be formulated as:

\vspace{-1.5em}
\small
\begin{equation}\label{ot_dist}
    d_{\text{OT}}(\mu(\bm{w}, \bm{t}), \mu(\bm{w}', \bm{u})) = \min_{\bm{P}} \sum_{i=1}^N \sum_{j=1}^M C_{i,j} P_{i,j},
\end{equation}
\normalsize
\vspace{-0.5em}

\noindent where $\bm{P} \in \mathbb{R}^{N \times M}_{\geq 0}$ is the transport plan, with entry $P_{i,j}$ denoting the amount of probability mass transported from $t_i$ to $u_j$; $\bm{C} \in \mathbb{R}^{N \times M}_{\geq 0}$ is the cost matrix, where $C_{i,j}$ represents the cost of transporting mass between word $w_i$ and $w'_j$.

The cost matrix $\bm{C}$ is constructed using the cosine distance between pre-trained word embeddings $\mathcal{E}^{\bm{w}}:=\{\bm{e}^{w_1},\bm{e}^{w_2},\ldots,\bm{e}^{w_{N}}\}$ (for the original topic words) and $\mathcal{E}^{\bm{w}'}:=\{\bm{e}^{w'_1},\bm{e}^{w'_2},\ldots,\bm{e}^{w'_{M}}\}$  (for the refined topic words). The cosine distance for each entry $C_{i,j}$ is computed as:

\vspace{-1.0em}
\begin{equation}\label{ot_cost}
   C_{i,j} := d_{\text{cos}}(\bm{e}^{w_i}, \bm{e}^{w'_j}), 
\end{equation}
\normalsize
\vspace{-1.5em}

\noindent where $d_{\text{cos}}(\bm{a}, \bm{b})$ denotes the cosine distance between the embedding vectors $\bm{a}$ and $\bm{b}$.

By minimizing this OT distance, the learned topic words from the NTM become aligned with the refined words suggested by the LLM, leading to more semantically coherent topics. This OT-based refinement loss is incorporated into the overall training objective, guiding the NTM to adjust its learned topics to match the LLM’s refined representations.

\subsection{Confidence-Weighted Topic Refinement}

LLMs, despite their powerful language capabilities, can sometimes produce hallucinated outputs—irrelevant or incorrect suggestions that do not align with the input data \citep{10.1145/3571730}. To mitigate the impact of such hallucinations, LLM-ITL introduces a confidence-weighted refinement mechanism.
The confidence mechanism assesses the reliability of the LLM’s refinements and adjusts their influence on the NTM’s training accordingly. This ensures that high-confidence refinements have a greater impact on the final topic representation, and vice versa.

We propose two methods for calculating topic labeling confidence of the LLM, considering whether the LLM is open-source or not: (1) Label token probability, applicable for open-source LLMs where the token probability of their generation is accessible; (2) Word intrusion confidence, available for both open and closed-source LLMs. 

\paragraph{Label Token Probability}
This method computes the product of the token probabilities for the topic label generated by the LLM. It reflects the LLM’s certainty in generating the specific topic label:

\vspace{-0.5em}
\small
\begin{equation}\label{uncertainty_token}
    \text{Conf}(\bm{w}^l)^{\text{prob}}:=\prod_{i=sol}^{eol}p(s_{i}|\bm{s}_{<i}, \bm{c}),
\end{equation}
\normalsize
where ``$sol$'' and ``$eol$'' denote the indices of ``start of label'' and ``end of label'' token, respectively; $p(s_i|\bm{s}_{<i},\bm{c})$ denotes the token probability of the $i$-th token; $\bm{c}$ denotes the input context to the LLM. 

\paragraph{Word Intrusion Confidence}  This method evaluates the proportion of irrelevant or ``intruder" words removed by the LLM during suggestion. A topic label generated based on a higher rate of intruder removal indicates that it is harder for the LLM to identify the topic from the original topic words, leading to lower confidence:

\vspace{-0.5em}
\small
\begin{equation}\label{uncertainty_intrusion}
    \text{Conf}(\bm{w}^l)^{\text{intrusion}}:=1-\frac{N^{\text{intruder}}}{N^{\bm{w}}},
\end{equation}
\normalsize
where $N^{\bm{w}}$ denotes the number of words in the given topic; $N^{\text{intruder}}$ denotes the number of intruders identified by the LLM. 

By incorporating the topic labeling confidence (Eq. \ref{uncertainty_token} or Eq. \ref{uncertainty_intrusion}) as a weight for the topic alignment loss, we adaptively adjust the impact of the LLM’s suggestion based on the confidence score. We write our confidence-weighted topic refinement objective as follows:

\vspace{-1em}
\small
\begin{equation}\label{reg_loss}
\min_{\bm{\phi}}\sum_{k=1}^{K}\text{Conf}(\bm{w}^l_k) \ d_{\text{OT}}(\mu(\bm{w}_k, \bm{t}_k),\mu(\bm{w}'_{k}, \bm{u}_k)). 
\end{equation}
\normalsize

\subsection{Integration with Warm-Up}
While the LLM's refinement can lead to more coherent topics, over-reliance on it may bias the topics toward the LLM's knowledge rather than reflecting the global information of the input corpus, thereby harming the topical representation of documents. To address this issue, we propose integrating the NTM with the LLM after a warm-up stage. This ensures that the refinement process begins only after the NTM has learned a stable topic representation, allowing the model to capture the core structure of the corpus before fine-tuning the topics with LLM guidance. By integrating the topic refinement objective with the training of an NTM with warm-up, we obtain the overall objective of LLM-ITL:

\vspace{-0.5em}
\small
\begin{equation}
\min_{\bm{\Theta}}(\mathcal{L}^{\text{ntm}}+ \gamma \cdot \mathbf{I}(t>T^\text{refine})   \cdot \mathcal{L}^{\text{refine}}),
\end{equation}
\normalsize
where $\bm{\Theta}:=\{\bm{\theta}, \bm{\phi}\}$ denotes model parameters; $\mathcal{L}^{\text{ntm}}$ and $\mathcal{L}^{\text{refine}}$ denote the NTM loss and refinement loss in Eq. \ref{ntm_loss} and Eq. \ref{reg_loss}, respectively; $\gamma$ controls the strength of focusing on the LLM's refinements; $t$ and $T^{\text{refine}}$ denote the current training step and the start of topic refinement, respectively; $\mathbf{I}(\cdot)$ denotes the indicator function. We provide the algorithm of LLM-ITL in Appendix \ref{algorithm}.


\section{Related Work}
\paragraph{Topic Models}
Classical topic models, such as Latent Dirichlet Allocation (LDA) \citep{10.5555/944919.944937} and its variants \citep{10.1145/1143844.1143859,10.5555/1036843.1036902,10.1145/2488388.2488514}, are Bayesian probabilistic models with various generative assumptions about the documents. Neural Topic Models (NTMs) \citep{10.5555/3305890.3305930,srivastava2017autoencoding,card-etal-2018-neural,dieng-etal-2020-topic,bianchi-etal-2021-pre,zhao2021neural,10.5555/3540261.3541177,NEURIPS2021_7b6982e5,pmlr-v139-duan21b,10.5555/3600270.3601909,xu-etal-2023-detime,xu-etal-2023-vontss,yang2023towards,miyamoto-etal-2023-dynamic,10.1609/aaai.v38i17.29895,chen2025stereographic,chen2025neural} use deep neural networks to learn topics and document representations. Clustering-based topic models \citep{sia-etal-2020-tired,grootendorst2022bertopic,angelov-inkpen-2024-topic} discover topics using clustering algorithms based on embeddings from pre-trained language models. Ours is not a specific topic model, but rather a general framework designed to integrate LLMs with a wide range of NTMs.

\paragraph{LLMs in Topic Modeling}
LLMs have been involved in topic modeling in various ways. \citet{6fe80f19eed94ff2b92c510b6e2670e8} investigate the use of ChatGPT \citep{openai2023chatgpt} to generate descriptions for topic words and found the effectiveness of these topic descriptions. Recent works leverage LLMs for topic model evaluation in different ways, such as applying LLMs for word intrusion or topic rating for topics \citep{rahimi-etal-2024-contextualized,stammbach-etal-2023-revisiting}, or keyword generation for documents \citep{10.1162/tacl_a_00744}. LLM-based topic models have emerged \citep{10386113,pham-etal-2024-topicgpt,mu-etal-2024-large,doi-etal-2024-topic}, which prompt LLMs to generate topics and assign topics to documents. Unlike existing methods that focus on document-level analysis relying on LLMs, our approach prompts LLMs to analyze a group of topic words, which are then used to guide the training of NTMs.
More recently, \citet{chang2024enhanced} show that LLMs are effective at refining topic words, leading to improved topic coherence. However, their method refines the topic words of trained topic models in a post-hoc manner, whereas ours incorporates refinement as a regularization term during training. Our method is also loosely related to uncertainty estimation of LLMs, and we omit the discussion on this in Appendix \ref{appendix_related_work}.

\section{Experiments}
\subsection{Experimental Setup}
\paragraph{Datasets}
We conduct experiments on four widely used and publicly available datasets for topic modeling, including 20Newsgroup \citep{10.5555/3091622.3091662} (\textbf{20News}), Reuters-21578 \citep{aletras-stevenson-2013-evaluating} (\textbf{R8}), \textbf{DBpedia} \citep{10.1007/978-3-540-76298-0_52} and \textbf{AGNews} \citep{10.5555/2969239.2969312}. Further details of these datasets are described in the Appendix \ref{appendix_data}. The number of mined topics (i.e., $K$) is commonly regarded as a hyper-parameter for the dataset \citep{zhao2021neural,wu2024survey}. For datasets containing long documents, such as 20News and R8, we set the number of topics to 50. For datasets with short documents, such as DBpedia and AGNews, we set the number to 25. We also run experiments at different $K$ values, which are reported in Appendix \ref{appendix:all_ks}.

\paragraph{Baselines}
The LLM-ITL framework is highly modular and can be seamlessly integrated with a wide range of NTMs. We integrated LLM-ITL with 8 commonly-used NTMs, as listed in Table \ref{tab:tc_main}. We also evaluate other types of topic models, including probabilistic models such as LDA \citep{10.5555/944919.944937}, cluster-based models like BERTopic \citep{grootendorst2022bertopic}, and the latest LLM-based model, TopicGPT \citep{pham-etal-2024-topicgpt}. Further details about these baselines and their settings are provided in Appendix \ref{appendix:baseline_setting}.

\paragraph{Settings of LLM-ITL}
We use LLAMA3-8B-Instruct \citep{dubey2024llama} in LLM-ITL for main experiments. For LLM generation, we use greedy decoding to enable deterministic output and set the maximum new generation tokens to 300. For OT computation, we use GloVe \citep{pennington-etal-2014-glove} word embeddings pre-trained on Wikipedia to construct the OT cost matrix, and compute the OT distance using the POT \citep{10.5555/3546258.3546336} package. For topic labeling confidence, we use label token probability in Eq. \ref{uncertainty_token} for our main experiments. For hyper-parameters of LLM-ITL, we maintain consistent settings across all datasets and base models it integrates with: the topic refinement strength $\gamma$ is set to 200; the number of words for the topic label is set to 2 when prompting the LLM; the warm-up steps are set as $T^{\text{refine}} = T^{\text{total}} - 50$, where $T^{\text{total}}$ denotes the total training steps of the base model without integration. Each component and hyper-parameter of LLM-ITL is studied in the following sections. Each trial\footnote{All experiments are conducted five times with different model random seeds throughout this work. The mean and standard deviation values of performance are reported.} of LLM-ITL in our experiment takes a few hours on a single 80GB A100 GPU.

\begin{table*}[t]
    \centering
    \resizebox{\textwidth}{!}{
    \begin{tabular}{lrr|rr|rr|rr}
    \toprule
    \multirow{2}{*}{\textbf{Model}} & \multicolumn{2}{c}{\textbf{20News}} & \multicolumn{2}{c}{\textbf{R8}} & \multicolumn{2}{c}{\textbf{DBpedia}} & \multicolumn{2}{c}{\textbf{AGNews}}\\
    \cmidrule(l){2-3} \cmidrule(l){4-5} \cmidrule(l){6-7} \cmidrule(l){8-9} 
    & $\bm{C_V}$ & \textbf{PN}  & $\bm{C_V}$ & \textbf{PN} & $\bm{C_V}$ & \textbf{PN} & $\bm{C_V}$ & \textbf{PN}\\
    \midrule
       LDA \citep{10.5555/944919.944937}  & 0.529\small{ ± 0.006} & 0.489\small{ ± 0.009} & 0.384\small{ ± 0.006} & 0.700\small{ ± 0.005} & 0.537\small{ ± 0.010} & 0.762\small{ ± 0.013} & 0.544\small{ ± 0.013} & 0.596\small{ ± 0.008}\\
       BERTopic \citep{grootendorst2022bertopic} & 0.404\small{ ± 0.005} & 0.342\small{ ± 0.008} & 0.323\small{ ± 0.010} & 0.697\small{ ± 0.001} & 0.545\small{ ± 0.015} & 0.720\small{ ± 0.009} & 0.506\small{ ± 0.016} & 0.450\small{ ± 0.010}\\
       TopicGPT \citep{pham-etal-2024-topicgpt} & NA & 0.363\small{ ± 0.000} & NA & 0.410\small{ ± 0.000}  & NA & 0.706\small{ ± 0.000} & NA &  0.634\small{ ± 0.000} \\
       \cmidrule(l){1-9}
       NVDM \citep{10.5555/3305890.3305930} &  0.261\small{ ± 0.008} & 0.141\small{ ± 0.006} & 0.298\small{ ± 0.010} & 0.359\small{ ± 0.010} & 0.329\small{ ± 0.017} & 0.175\small{ ± 0.009} & 0.355\small{ ± 0.010} & 0.252\small{ ± 0.009}\\
       + LLM-ITL & 0.336\small{ ± 0.014} & 0.138\small{ ± 0.007} & 0.374\small{ ± 0.016} & 0.360\small{ ± 0.011} & 0.479\small{ ± 0.025} & 0.170\small{ ± 0.006} & 0.455\small{ ± 0.016} & 0.248\small{ ± 0.013} \\
       & \bm{$\uparrow$} \textbf{28.7\%} & $\downarrow$ 2.1\% & \bm{$\uparrow$} \textbf{25.5\%} & \bm{$\uparrow$} \textbf{0.3\%} & \bm{$\uparrow$} \textbf{45.6\%} & $\downarrow$ 2.9\% & \bm{$\uparrow$} \textbf{28.2\%} & $\downarrow$ 1.6\%  \\
       \cmidrule(l){1-9}
       PLDA \citep{srivastava2017autoencoding} & 0.368\small{ ± 0.011} & 0.272\small{ ± 0.008} & 0.375\small{ ± 0.004} & 0.525\small{ ± 0.006} & 0.502\small{ ± 0.014} & 0.664\small{ ± 0.009} & 0.551\small{ ± 0.012} & 0.481\small{ ± 0.010}\\
       + LLM-ITL & 0.525\small{ ± 0.010} & 0.272\small{ ± 0.011} & 0.466\small{ ± 0.011} & 0.522\small{ ± 0.008} & 0.627\small{ ± 0.020} & 0.660\small{ ± 0.005} & 0.625\small{ ± 0.017} & 0.486\small{ ± 0.008}\\
       & \bm{$\uparrow$} \textbf{42.7\%} & \bm{$\uparrow$} \textbf{0.0\%} & \bm{$\uparrow$} \textbf{24.3\%} & $\downarrow$ 0.6\% & \bm{$\uparrow$} \textbf{24.9\%} & $\downarrow$ 0.6\% & \bm{$\uparrow$} \textbf{13.4\%} & \bm{$\uparrow$} \textbf{1.0\%} \\
       \cmidrule(l){1-9}
       SCHOLAR \citep{card-etal-2018-neural} & 0.479\small{ ± 0.019} & 0.582\small{ ± 0.010} & 0.386\small{ ± 0.004} & 0.680\small{ ± 0.013} & 0.608\small{ ± 0.019} & 0.825\small{ ± 0.015} & 0.576\small{ ± 0.011} & 0.638\small{ ± 0.003}\\ 
       + LLM-ITL & 0.591\small{ ± 0.012} & 0.568\small{ ± 0.010} & 0.439\small{ ± 0.008} & 0.680\small{ ± 0.012} & 0.678\small{ ± 0.017} & 0.828\small{ ± 0.013} & 0.655\small{ ± 0.012} & 0.639\small{ ± 0.002}\\
       & \bm{$\uparrow$} \textbf{23.4\%} & $\downarrow$ 2.4\% & \bm{$\uparrow$} \textbf{13.7\%} & \bm{$\uparrow$} \textbf{0.0\%} & \bm{$\uparrow$} \textbf{11.5\%} & \bm{$\uparrow$} \textbf{0.4\%} & \bm{$\uparrow$} \textbf{13.7\%} & \bm{$\uparrow$} \textbf{0.2\%} \\
       \cmidrule(l){1-9}
       ETM \citep{dieng-etal-2020-topic} & 0.491\small{ ± 0.006} & 0.404\small{ ± 0.010} & 0.433\small{ ± 0.006} & 0.679\small{ ± 0.018} & 0.513\small{ ± 0.004} & 0.762\small{ ± 0.011} & 0.534\small{ ± 0.013} & 0.568\small{ ± 0.008}\\
       + LLM-ITL & 0.578\small{ ± 0.008} & 0.398\small{ ± 0.010} & 0.571\small{ ± 0.010} & 0.686\small{ ± 0.012} & 0.704\small{ ± 0.015} & 0.742\small{ ± 0.016} & 0.644\small{ ± 0.017} & 0.569\small{ ± 0.005}\\
       & \bm{$\uparrow$} \textbf{17.7\%} & $\downarrow$ 1.5\% & \bm{$\uparrow$} \textbf{31.9\%} & \bm{$\uparrow$} \textbf{1.0\%} & \bm{$\uparrow$} \textbf{37.2\%} & $\downarrow$ 2.6\% & \bm{$\uparrow$} \textbf{20.6\%} & \bm{$\uparrow$} \textbf{0.2\%} \\
       \cmidrule(l){1-9}
        NSTM \citep{zhao2021neural} & 0.444\small{ ± 0.015} & 0.373\small{ ± 0.005} & 0.412\small{ ± 0.004} & 0.665\small{ ± 0.014} & 0.652\small{ ± 0.005} & 0.767\small{ ± 0.011} & 0.588\small{ ± 0.022} & 0.593\small{ ± 0.010}\\
        + LLM-ITL & 0.521\small{ ± 0.015} & 0.360\small{ ± 0.006} & 0.549\small{ ± 0.011} & 0.673\small{ ± 0.006} & 0.700\small{ ± 0.013} & 0.752\small{ ± 0.014} & 0.675\small{ ± 0.012} & 0.585\small{ ± 0.010}\\
        & \bm{$\uparrow$} \textbf{17.3\%} & $\downarrow$ 3.5\% & \bm{$\uparrow$} \textbf{33.3\%} & \bm{$\uparrow$} \textbf{1.2\%} & \bm{$\uparrow$} \textbf{7.4\%} & $\downarrow$ 2.0\% & \bm{$\uparrow$} \textbf{14.8\%} & $\downarrow$ 1.3\% \\
       \cmidrule(l){1-9}
       CLNTM \citep{10.5555/3540261.3541177} & 0.490\small{ ± 0.014} & 0.575\small{ ± 0.011} & 0.361\small{ ± 0.008} & 0.691\small{ ± 0.005} & 0.500\small{ ± 0.015} & 0.683\small{ ± 0.040} & 0.558\small{ ± 0.028} & 0.607\small{ ± 0.014}\\
       + LLM-ITL & 0.612\small{ ± 0.010} & 0.576\small{ ± 0.005} & 0.436\small{ ± 0.008} & 0.691\small{ ± 0.005} & 0.612\small{ ± 0.019} & 0.684\small{ ± 0.039} & 0.655\small{ ± 0.020} & 0.594\small{ ± 0.011}\\
       & \bm{$\uparrow$} \textbf{24.9\%} & \bm{$\uparrow$} \textbf{0.2\%} & \bm{$\uparrow$} \textbf{20.8\%} & \bm{$\uparrow$} \textbf{0.0\%} & \bm{$\uparrow$} \textbf{22.4\%} & \bm{$\uparrow$} \textbf{0.1\%} & \bm{$\uparrow$} \textbf{17.4\%} & $\downarrow$ 2.1\% \\
       \cmidrule(l){1-9}
       WeTe \citep{wang2022representing} & 0.495\small{ ± 0.017} & 0.318\small{ ± 0.009} & 0.529\small{ ± 0.011} & 0.649\small{ ± 0.002} & 0.546\small{ ± 0.003} & 0.721\small{ ± 0.012} & 0.560\small{ ± 0.006} & 0.516\small{ ± 0.004}\\
       + LLM-ITL &  0.583\small{ ± 0.028} & 0.316\small{ ± 0.008} & 0.599\small{ ± 0.015} & 0.656\small{ ± 0.005} & 0.613\small{ ± 0.015} & 0.703\small{ ± 0.012} & 0.623\small{ ± 0.007} & 0.517\small{ ± 0.004}\\
       & \bm{$\uparrow$} \textbf{17.8\%} & $\downarrow$ 0.6\% & \bm{$\uparrow$} \textbf{13.2\%} & \bm{$\uparrow$} \textbf{1.1\%} & \bm{$\uparrow$} \textbf{12.3\%} & $\downarrow$ 2.5\% & \bm{$\uparrow$} \textbf{11.2\%} & \bm{$\uparrow$} \textbf{0.2\%} \\
       \cmidrule(l){1-9}
       ECRTM \citep{pmlr-v202-wu23c} & 0.323\small{ ± 0.014} & 0.516\small{ ± 0.005} & 0.308\small{ ± 0.006} & 0.673\small{ ± 0.008} & 0.581\small{ ± 0.026} & 0.668\small{ ± 0.202} & 0.438\small{ ± 0.035} & 0.554\small{ ± 0.019}\\
       + LLM-ITL & 0.551\small{ ± 0.025} & 0.519\small{ ± 0.002} & 0.364\small{ ± 0.010} & 0.667\small{ ± 0.007} & 0.684\small{ ± 0.013} & 0.696\small{ ± 0.142} & 0.505\small{ ± 0.025} & 0.553\small{ ± 0.020} \\
       & \bm{$\uparrow$} \textbf{70.6\%} & \bm{$\uparrow$} \textbf{0.6\%} & \bm{$\uparrow$} \textbf{18.2\%} & $\downarrow$ 0.9\% & \bm{$\uparrow$} \textbf{17.7\%} & \bm{$\uparrow$} \textbf{4.2\%} & \bm{$\uparrow$} \textbf{15.3\%} & $\downarrow$ 0.2\% \\
       \bottomrule
    \end{tabular}
    }
    \caption{Topic coherence ($C_V$) and topic alignment (PN). ``NA'' indicates the evaluation is not applicable. The performance improvement of LLM-ITL over its base model is computed.}\label{tab:tc_main}
\end{table*}

\paragraph{Evaluation Metrics}
We evaluate both the topic quality and the document representation quality for topic models. For topic quality, we apply the widely used \textbf{topic coherence} metric, $\bm{C_V}$ \citep{10.1145/2684822.2685324}. We report the average $C_V$ values of all learned topics.  For document representation quality, we evaluate the alignment between a document’s true label and the top-weighted topic of its topical representation, known as \textbf{topic alignment} \citep{10.5555/3042817.3043005,pham-etal-2024-topicgpt}. This is commonly evaluated using external clustering metrics, Purity and Normalized Mutual Information (NMI). Since Purity and NMI are considered equally important, fall within the same range (from 0 to 1), and are often reported together, we report their average as \textbf{PN}, serving as an overall indicator of topic alignment performance. Detailed results for Purity and NMI are provided in Appendix \ref{purity_nmi}. Intuitively, topic coherence ($C_V$) reflects how coherent the learned topic words are, while topic alignment (PN) indicates how well the model represents the documents through the learned topics. Further details on the calculation of these metrics are provided in Appendix \ref{appendix:eval_metrics}. In addition, we compute the performance improvement of LLM-ITL over its base model as: $\frac{S(\text{base+LLM-ITL}) - S(\text{base})}{S(\text{base})}$, where $S(\cdot)$ represents the evaluation metric. As for other topic model evaluation metrics, including topic diversity (TD) \citep{dieng-etal-2020-topic} and overall topic quality (TQ) \citep{wang2022representing} are reported in Appendix \ref{appendix:TD} and \ref{appendix:TQ}, respectively.

\begin{table*}[t]
\centering
\resizebox{\textwidth}{!}{
\begin{tabular}{p{5.5cm}cp{11.5cm}c}
\toprule
\textbf{Document} & \textbf{Model} & \textbf{Topic} & \textbf{Proportion} \\
\midrule
\multirow{4}{*}{\pbox{5.5cm}{\vspace{2mm} Here's a listing that I came accross a while ago.  This question seems to 
come up often enough that I figured this would be of interest.  Note that the server ``X Appeal'' for DOS is available in demo form on the internet via anonymous ftp. This is one way of quickly checking out the feasability of 
using your system as an X server.  Enjoy! - Pete 
\\ \textit{*** Many words omitted here ***} \\ 
1280x960x1 (TT, SM194) color 320x200x4 color 640x200x2 color 640x480x4 color 320x480x8
Ethernet Card: Atari Card (Mega or VME bus) Riebl/Wacker (Mega or VME bus) ----- End Enclosure -----}} 
& \multirow{4}{*}{LDA} & [1] window file program problem run work running machine time version & 0.23\\
& & [2] card driver monitor color video mode vga window screen problem & 0.19\\
& & [3] image data graphic package software program format tool file processing & 0.16\\
& & [4] mb mac mhz bit chip card scsi ram cpu memory & 0.10\\
& & ... & ...\\
\cmidrule(l){2-4}
& \multirow{5}{*}{TopicGPT} & [1] Software Development -- The document provides a list of X servers that can be used on non-UNIX networked machines. & \multirow{2}{*}{NA}\\
& & [2] Internet Culture -- The document mentions the availability of X servers for various operating systems and provides information on how to access the file via anonymous ftp. &  \multirow{3}{*}{NA}\\
\cmidrule(l){2-4}
& \multirow{7}{*}{LLM-ITL} & [1] Software Development -- application model version program software designed tool development implementation window & \multirow{2}{*}{0.23}\\
& & [2] Computer Hardware -- computer bios disk pc chip controller ram memory card apple & \multirow{2}{*}{0.13}\\
& & [3] Service Support -- support provide offer access available providing includes provides use feature & \multirow{2}{*}{0.11}\\
& & ... & ... \\
\bottomrule
\end{tabular}
}
\caption{Examples of different topic models' output for a given document from 20News. Only the document's top assigned/weighted ($>= 0.1$) topics of its topical proportion/representation are listed in LDA and LLM-ITL (with ETM as the base model).}\label{qualitive}
\end{table*}

\subsection{Results}
\paragraph{Topic Coherence and Alignment}
We show the performance of topic coherence and topic alignment for different models in Table \ref{tab:tc_main}. We summarize the following remarks based on the results: (1) For topic coherence, using LLM-ITL significantly improves performance across all cases, with minimum gains of +7.4\% (NSTM as the base model on DBpedia) and maximum gains of +70.6\% (ECRTM as the base model on 20News) over the base model. (2) In terms of topic alignment, integrating LLM-ITL demonstrates comparable performance to its base model, with changes ranging from -3.5\% to +4.2\%. (3) Moreover, for long-document corpora such as 20News and R8, applying LLM-ITL outperforms existing LLM-based topic models like TopicGPT in terms of topic alignment in most cases. This suggests that LLMs alone struggle to fully capture the topics of long documents, highlighting the advantages of integrating NTMs to enhance topical representation for document collections. Additional results with different settings and evaluation metrics are presented in Appendix \ref{appendix:all_ks} and Appendix \ref{sec:extra_coherence}.



\paragraph{Qualitative Analysis} During the inference phase, LLM-ITL infers the topic proportion for a given document from the NTM component, and obtains the topic label from the LLM component, as shown in Table \ref{qualitive}. We can observe that (1) Compared to topic models with top-words topics such as LDA, LLM-ITL provides more coherent topic words and offers topic labels, making the semantic meaning of the topics easier to identify. (2) Compared to the LLM-based topic model TopicGPT, LLM-ITL can obtain topic proportions as an indicator of the importance or relevance of topics to the document, offering more practical usage. For example, for TopicGPT, ``Internet Culture'' should be less relevant for the example document than ``Software Development'' if a good topic proportion is available.

\begin{figure}[t]
    \centering
    \includegraphics[width=0.48\textwidth]{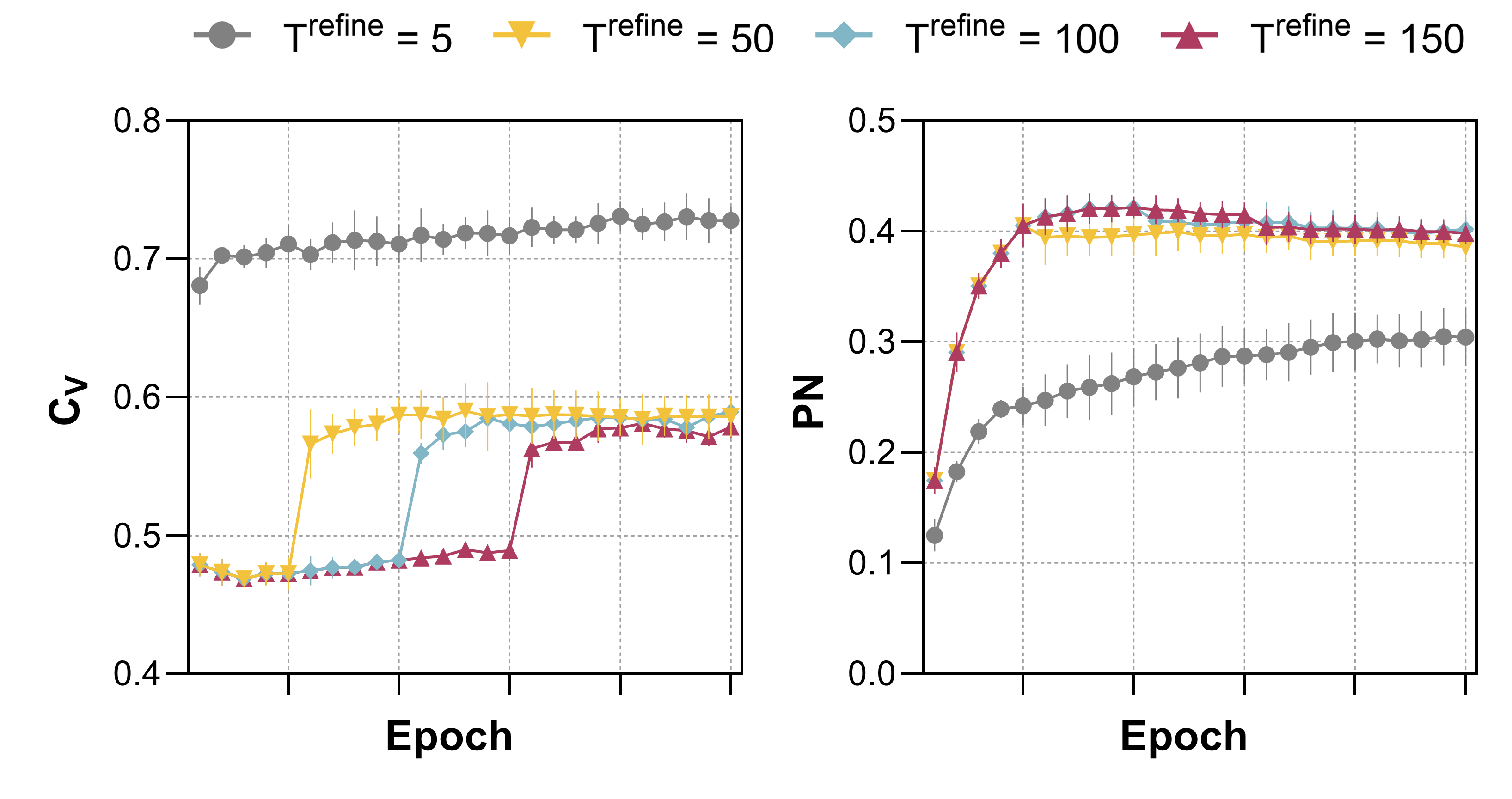}
  \caption{Learning curves of LLM-ITL (ETM as the base model) with different $T^{\text{refine}}$ in terms of topic coherence ($C_V$) and topic alignment (PN) on 20News.}
  \label{fig:hyper_warm}
\end{figure}

\paragraph{Balancing Topic Coherence and Alignment} \label{sec:warm} 
Over-reliance on the LLM's refinement may introduce out-of-corpus information, thereby harming topical representation of documents for the corpus (i.e., topic alignment). Here, we demonstrate the effectiveness of using warm-up integration to balance topic coherence and alignment. From the learning curves illustrated in Figure \ref{fig:hyper_warm}, we observe that starting topic refinement earlier (e.g., $T^\text{refine} = 5$) can result in greater improvements in topic coherence but leads to poorer topic alignment performance. For larger values of $T^\text{refine}$, the performance in terms of both metrics becomes better balanced and comparable, indicating the effectiveness of warm-up integration and low sensitivity within a certain range. For more hyper-parameter studies of LLM-ITL, see Appendix \ref{appendix:hyper}.

\begin{figure}[ht]
    \centering
    \includegraphics[width=0.48\textwidth]{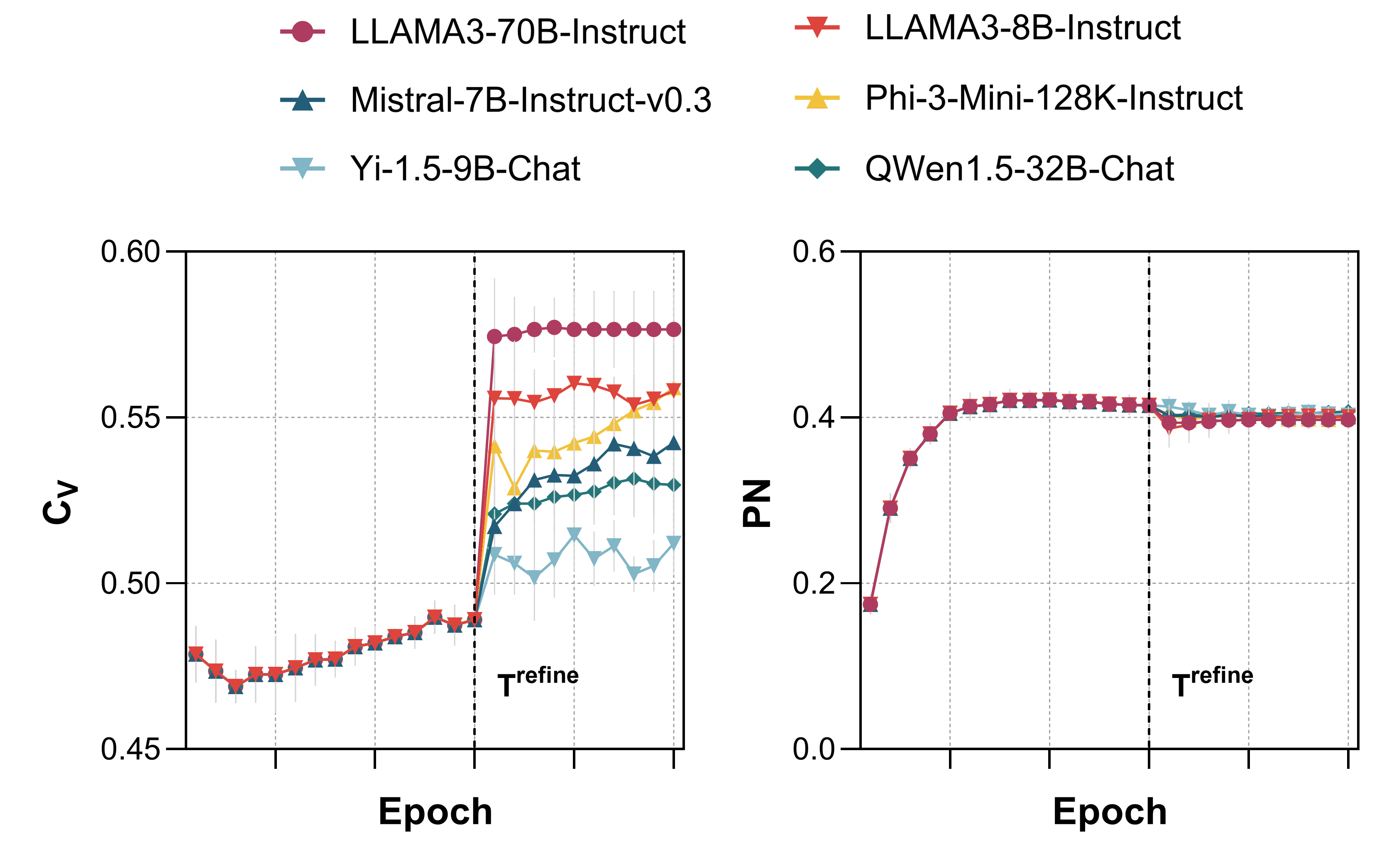}
  \caption{Learning curves of LLM-ITL (ETM as the base model) with different LLMs in terms of topic coherence ($C_V$) and topic alignment (PN) on 20News.}\label{fig:diff_llms}
\end{figure}

\paragraph{Flexibility with different LLMs} \label{other_llms}
LLM-ITL is a framework compatible with most LLMs. Here, we examine the flexibility of LLM-ITL by integrating it with various LLMs. Apart from LLAMA3-8B-Instruct \citep{dubey2024llama}, we implement LLM-ITL with the latest open-sourced LLMs, including Mistral-7B-Instruct-v0.3 \citep{jiang2023mistral}, Phi-3-Mini-128K-Instruct \citep{abdin2024phi}, Yi-1.5-9B-Chat \citep{young2024yi}, Qwen1.5-32B-Chat \citep{qwen} and LLAMA3-70B-Instruct \citep{dubey2024llama}. As shown in Figure \ref{fig:diff_llms}, LLM-ITL consistently improves topic coherence of its base model across different LLMs, and the improvement can be further enhanced when using larger LLMs such as LLAMA3-70B-Instruct.

\begin{figure}[t]
    \centering
    \includegraphics[width=0.48\textwidth]{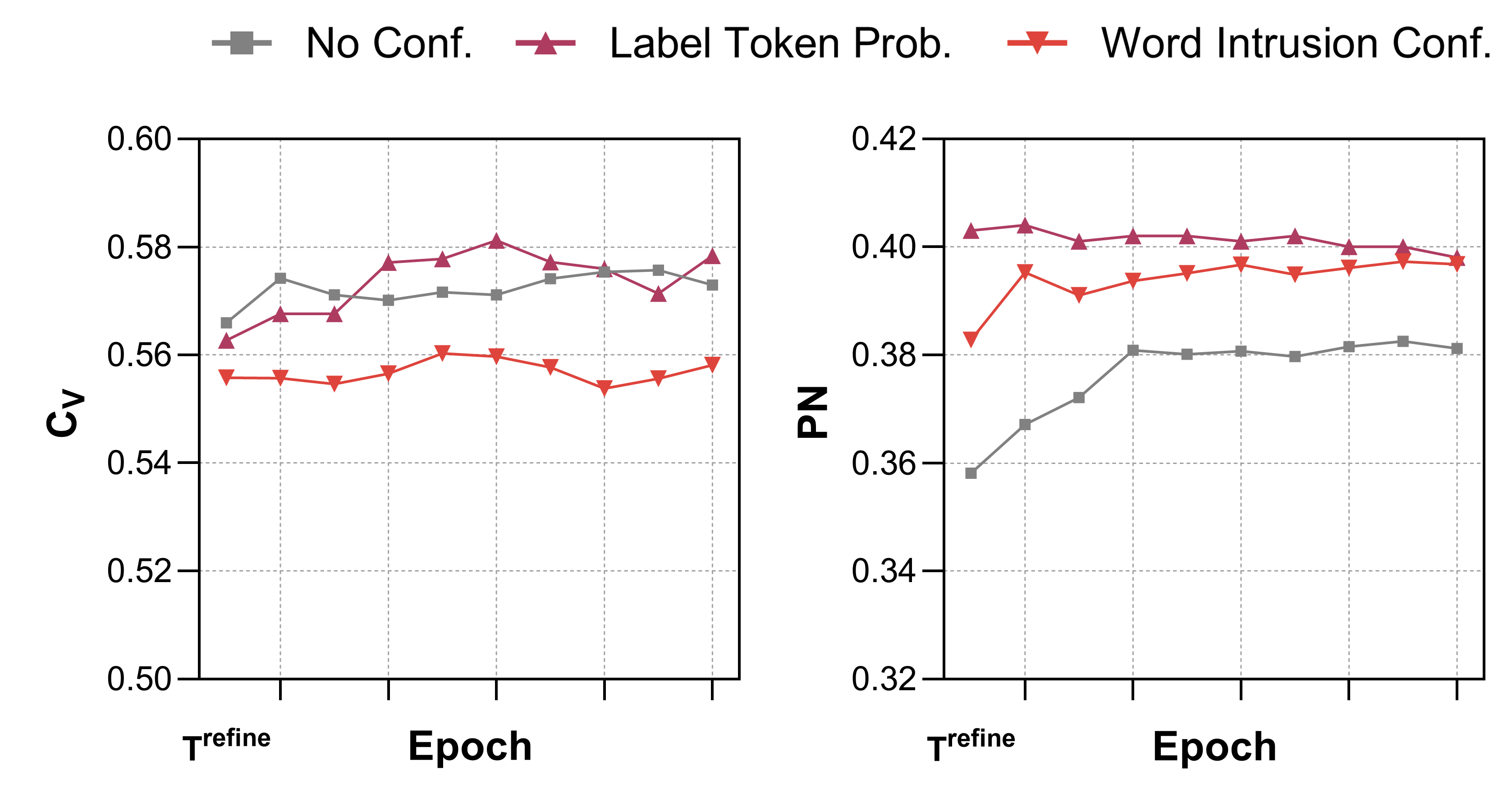}
  \small{\caption{Ablation for confidence. Error bars are omitted for clarity in the figure.}\label{fig:ablation}}
\end{figure}

\paragraph{Ablation Study for Confidence}\label{sec:ablation_conf}
Here, we investigate the effectiveness of including the confidence scores during topic refinement. As illustrated in Figure \ref{fig:ablation}, label token probability and word intrusion confidence consistently yield better performance in terms of PN. This suggests that by including proposed confidence during topic refinement, we reduce the impact of potential noisy suggestions from the LLM and achieve better topical representation for documents. For further studies on alternative LLM confidence measures, see Appendix \ref{appendix:conf}.

\paragraph{Ablation Study of OT}

\begin{table}[t]
    \centering
    \begin{tabular}{ccc}
    \toprule
    & $\bm{C_V}$ & \textbf{PN} \\
    \midrule
      KL & 0.480\small{ ± 0.007} & 0.402\small{ ± 0.012}\\
      JSD & 0.479\small{ ± 0.003} & 0.403\small{ ± 0.011}\\
      HD & 0.480\small{ ± 0.011} & 0.403\small{ ± 0.012}\\
      TVD & 0.480\small{ ± 0.008} & \textbf{0.405}\small{ ± 0.011}\\
      OT & \textbf{0.578}\small{ ± 0.008} & 0.398\small{ ± 0.010}\\
       \bottomrule
    \end{tabular}
\caption{Ablation study of OT. The best performance is highlighted in boldface for both metrics.}
\label{tab:ot_ablation}
\end{table}
Here, we conducted an ablation study on our OT-based topic refinement approach. In this study, we apply different metrics to measure the differences between the topic word distributions generated by the NTM and those refined by the LLM. Specifically, we explored alternative metrics to optimal transport (\textbf{OT}) within the LLM-ITL framework, including Kullback–Leibler (\textbf{KL}) divergence, Jensen–Shannon divergence (\textbf{JSD}), Hellinger distance (\textbf{HD}), and Total Variation distance (\textbf{TVD}). 

As shown in Table \ref{tab:ot_ablation}, which presents results from experiments on 20News with $K=50$ using ETM as the base model, our OT-based approach demonstrates a significant advantage in improving topic coherence compared to alternative distribution measurement methods.

\section{Conclusion}
In this paper, we introduced LLM-ITL, a novel framework that integrates Large Language Models (LLMs) with Neural Topic Models (NTMs) to address the limitations of both traditional topic models and the direct use of LLMs for topic discovery. By incorporating a confidence-weighted Optimal Transport (OT)-based topic alignment, LLM-ITL improves the interpretability and coherence of topics while maintaining the quality of document representations. Our framework effectively leverages the strengths of both LLMs and NTMs, offering a flexible, scalable, and efficient solution for topic modeling. Extensive experiments on benchmark datasets demonstrate that applying LLM-ITL with NTMs significantly boosts topic interpretability while maintaining document representation quality.

\section{Limitation}
The proposed framework relies on the refinements generated by the LLM. 
Over-reliance on the LLM during the refinement process may introduce out-of-corpus information or bias the topics toward the LLM's pre-training knowledge. This can negatively impact the alignment of learned topics with the input corpus, particularly in cases where the corpus diverges significantly from the LLM's training data.


\section*{Acknowledgment}
We thank the anonymous (meta) reviewers in ARR for their valuable feedback, which has significantly strengthened this work.

\bibliography{custom}

\onecolumn
\newpage

\appendix

\section{Prompts}

\setcounter{figure}{0}
\renewcommand{\thefigure}{A\arabic{figure}}

\begin{figure*}[!h]
    \centering
    \includegraphics[width=0.95\textwidth]{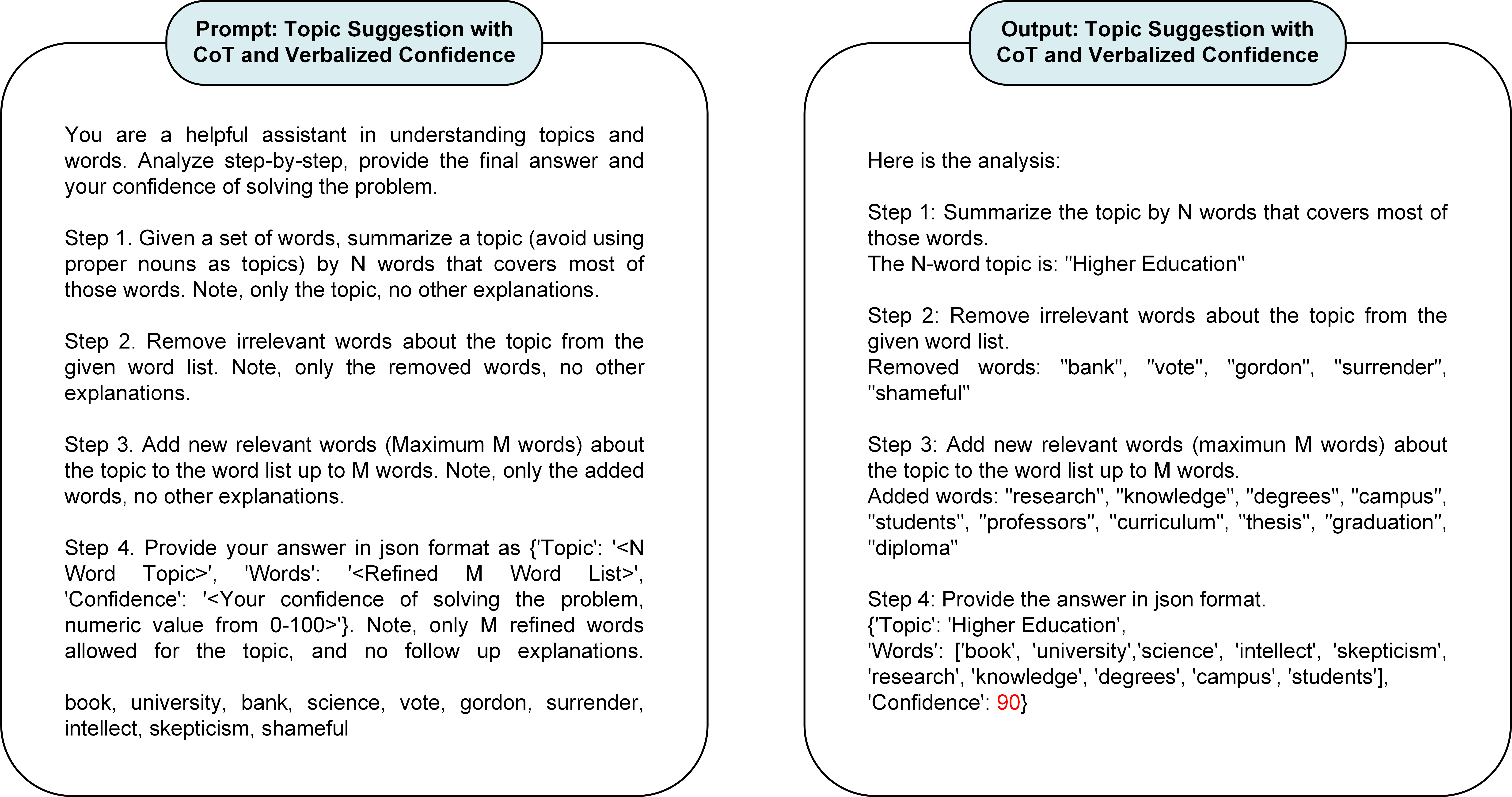}
  \caption{Prompt and output of topic suggestion with CoT and verbalized confidence. The number in \textbf{red} color in the LLM's output represents the verbalized confidence. $N=2$ and $M=10$ in this example.}
  \label{fig:prompt_refine_conf}
\end{figure*}

\begin{figure*}[!h]
    \centering
    \includegraphics[width=0.95\textwidth]{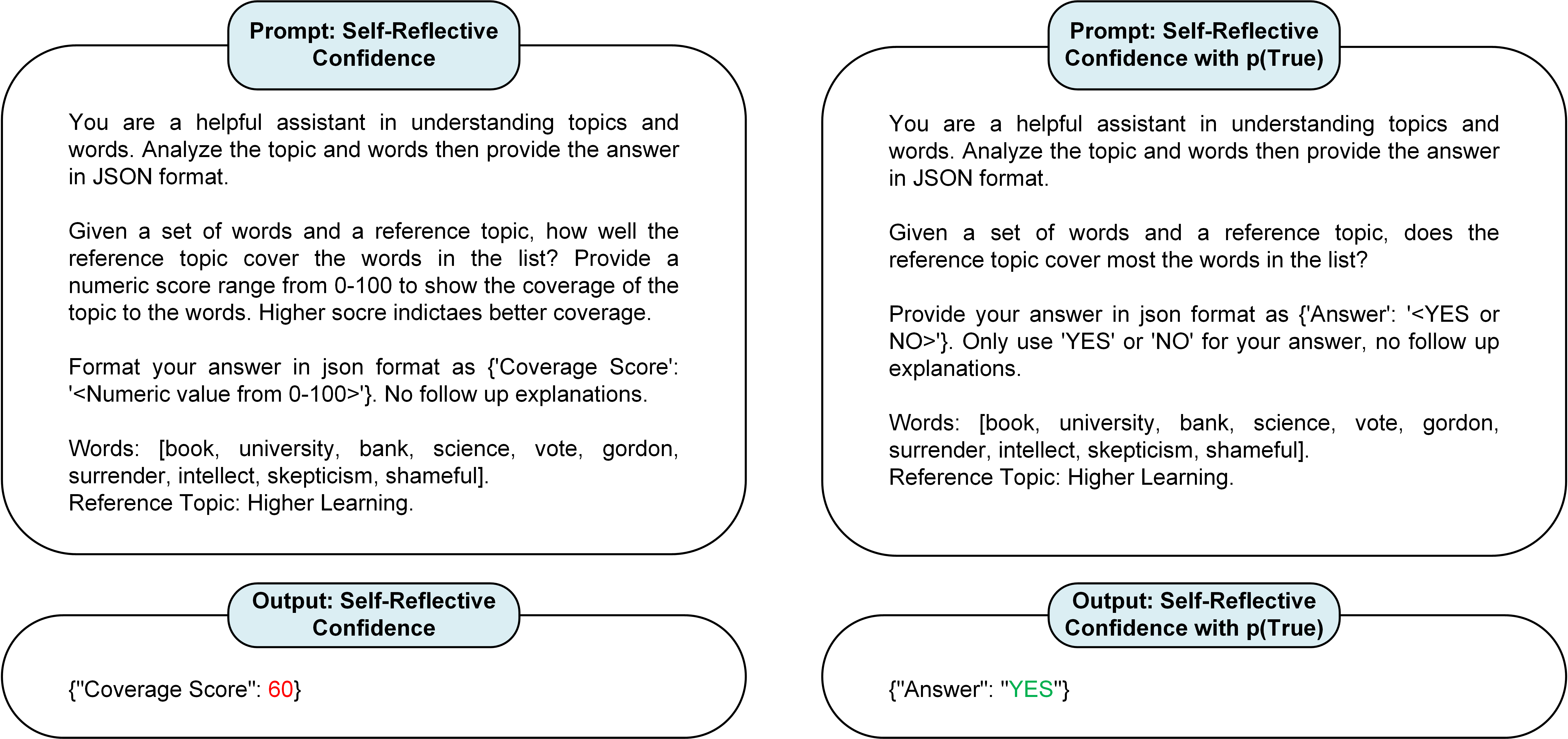}
  \caption{Prompt and output of self-reflective confidence and p(True). \textbf{Left}: self-reflective confidence where the number in \textbf{red} color represents the confidence. \textbf{Right}: p(True) confidence where the token probability of ``YES'' (p) in \textbf{green} color or ``NO'' (1-p) is used as confidence.}
  \label{fig:prompt_self_reflective}
\end{figure*}

\newpage

\section{Algorithm}\label{algorithm}
\begin{algorithm}[H]
\small
\SetAlgoLined
\caption{ Algorithm for LLM-ITL }
\KwIn{Train documents; An LLM; pre-training word embeddings; Hyper-parameters $T^{\text{refine}}$, $\gamma$; Training iteration $I$; Number of topics $K$.}
\textbf{Initialize:} {Initialize the parameters $\bm{\theta},\bm{\phi}$ of the NTM}.\\

\textbf{/*Warm-up*/}\\
\For{$i=1:T^{\text{refine}}$}{
Compute NTM loss by Eq. \ref{ntm_loss};\\
Compute gradients w.r.t $\bm{\theta}$ and $\bm{\phi}$;\\
Update $\bm{\theta}$ and $\bm{\phi}$ based on the gradients;
}

\textbf{/*Topic Refinement */}\\
\For{$i=T^{\text{refine}}:I$}{
    \For{$k=1:K$ }{
        Obtain topic distribution $\bm{t}_k$ by Eq. \ref{eq_t};\\
        Obtain topic words $\bm{w}_k$ by Eq. \ref{eq_w};\\
        Obtain refined words $\bm{w}_k^{'}$ from the LLM by Eq. \ref{llm_opt};\\
        Construct OT cost matrix by Eq. \ref{ot_cost};\\
        Compute OT distance by Eq. \ref{ot_dist};\\
        \If{Open-Source LLM}{Compute topic labeling confidence by Eq. \ref{uncertainty_token}}
        \Else{Compute topic labeling confidence by Eq. \ref{uncertainty_intrusion}}
    }
    Compute $\mathcal{L}^{\text{refine}}$ by Eq. \ref{reg_loss};\\
    Compute $\mathcal{L}^{\text{ntm}}$ by Eq. \ref{ntm_loss}; \\
    Compute overall loss by $\mathcal{L}^{\text{ntm}} + \gamma \cdot \mathcal{L}^{\text{refine}}$ ;\\
    Compute gradients w.r.t $\bm{\theta}$ and $\bm{\phi}$; \\
    Update $\bm{\theta}$ and $\bm{\phi}$ based on the gradients; \\
}
\KwOut{Trained NTM with $\bm{\theta},\bm{\phi}$.}
\end{algorithm}

\twocolumn

\section{Related work: LLM Uncertainty Estimation}\label{appendix_related_work}
Uncertainty estimation for LLMs \citep{geng-etal-2024-survey} is emerging with the rapid usage of LLMs and their risk of hallucination \citep{10.1145/3571730}. Sequence probability \citep{ren2023outofdistribution} leverages token probabilities to measure answer confidence. Verbalized confidence \citep{tian-etal-2023-just,xiong2024can} utilizes the LLM's own capability to evaluate its answer uncertainty. Consistency-based \citep{lin2024generating,manakul-etal-2023-selfcheckgpt} approaches sample multiple outputs from the LLM and measure answer consistency as uncertainty. Entropy-based \citep{kuhn2023semantic,pmlr-v235-hou24b} approaches estimate the output space from multiple LLM outputs and compute entropy as uncertainty for the answer. Hybrid frameworks \citep{chen-mueller-2024-quantifying,gao-etal-2024-spuq} combine different approaches for a comprehensive estimation. Internal states \citep{chen2024inside} are another useful source for LLM uncertainty quantification. Unlike those works that estimate uncertainty for LLMs in general natural language generation tasks, ours focuses on task-specific uncertainty of the LLM in suggesting topic words.

\section{Detailed Experimental Settings}

\subsection{Details of Dataset}\label{appendix_data}

\setcounter{table}{0} 
\renewcommand{\thetable}{D\arabic{table}}

\begin{table}[H]  
  \centering
  \resizebox{0.5\textwidth}{!}{
  \begin{tabular}{cccccc}
    \toprule
    Dataset & \# Docs Train & \# Docs Test & Voc Size &  Avg. Doc Length & \# Labels\\
    \midrule
    20News              & 11778 & 2944 & 13925 & 150 & 20 \\
    R8 & 5485 & 2189 & 5338 & 102 & 8 \\
    DBpedia           & 15598 & 3899   & 8550 & 51 & 14 \\
    AGNews & 16000 & 4000 & 8389 & 38 & 4\\
  \bottomrule
\end{tabular}
}
\caption{Statistics of the Datasets}\label{data_statisc}
\end{table}

We conduct experiments on 20News\footnote{https://huggingface.co/datasets/SetFit/20\_newsgroups}, R8\footnote{https://huggingface.co/datasets/yangwang825/reuters-21578}, DBpedia\footnote{https://huggingface.co/datasets/fancyzhx/dbpedia\_14}, and AGNews\footnote{https://huggingface.co/datasets/fancyzhx/ag\_news}. For DBpedia and AGNews, we randomly sample a subset of 20,000 documents. We retain the original text documents for models that accept text as input, and preprocess the documents into Bag-of-Words (BoW) format for models that are trained on BoWs. We convert the documents into BoW vectors through the following steps: First, we clean the documents by removing special characters and stop words, followed by tokenization. Next, we build the vocabulary by including words with a document frequency greater than five and less than 80\% of the total documents. Since we use the pre-training word embeddings of GloVe \citep{pennington-etal-2014-glove}, we further filter the vocabulary by retaining only the words that are in the GloVe vocabulary. Finally, we transform the documents into BoWs based on the filtered vocabulary set. The statistics of the preprocessed datasets are summarized in Table \ref{data_statisc}. 

\subsection{Details of Baselines} \label{appendix:baseline_setting}
We run the following topic models in our experiments, including Latent Dirichlet Allocation (\textbf{LDA}) \citep{10.5555/944919.944937}, the most popular probabilistic topic model that generates documents by mixtures of topics; Neural Variational Document Model (\textbf{NVDM}) \citep{10.5555/3305890.3305930}, a pioneering NTM based on the VAE framework; LDA with Products of Experts (\textbf{PLDA}) \citep{srivastava2017autoencoding}, an NTM that uses a product of experts instead of the mixture model
 in LDA; Neural Topic Model with Covariates, Supervision, and Sparsity (\textbf{SCHOLAR}) \citep{card-etal-2018-neural}, an NTM that leverages extra information from metadata; Embedded Topic Model (\textbf{ETM}) \citep{dieng-etal-2020-topic}, which involves
word and topic embeddings in the generative process of documents; Neural Sinkhorn Topic Model (\textbf{NSTM}) \citep{zhao2021neural}, an NTM based on an optimal transport framework; Contrastive Learning Neural Topic Model (\textbf{CLNTM}) \citep{10.5555/3540261.3541177}, an NTM that is based on the contrastive learning framework; \textbf{BERTopic} \citep{grootendorst2022bertopic}, a recent clustering-based topic model that utilizes embeddings from pre-trained language models; \textbf{WeTe} \citep{wang2022representing}, which represents a mixture of word embeddings as a mixture of topic embeddings; Embedding Clustering Regularization Topic Model (\textbf{ECRTM}) \citep{pmlr-v202-wu23c}, an NTM with embedding clustering regularization; \textbf{TopicGPT} \citep{pham-etal-2024-topicgpt}, a latest LLM-based topic model that leverages an LLM for topic generation and assignment. 

As for the implementations of baseline models, we use Mallet\footnote{https://radimrehurek.com/gensim\_3.8.3/models/wrappers/\\ldamallet.html} for LDA with Gibbs sampling, and the original implementations for the other models. For NTMs including NVDM, PLDA, SCHOLAR, ETM, NSTM, CLNTM, WeTe and ECRTM, we tune their hyper-parameters for the datasets; For BERTopic, we fine-tune the topic representations after the topics are learned, as suggested by their implementation\footnote{https://maartengr.github.io/BERTopic/index.html}. For TopicGPT, we use GPT-4 for topic generation and GPT-3.5 for topic assignment, randomly sampling 600 documents from the training set for each dataset, as suggested by their paper. We run all models except TopicGPT five times in each experiment and report the mean and standard deviation of their performance. For TopicGPT, we run it once for each experiment using a temperature value of zero to enable deterministic output, following the setting of their paper.

\subsection{Details of Evaluation Metrics} 
\label{appendix:eval_metrics}
For topic evaluation, we apply the commonly-used topic coherence metric, $\bm{C_V}$ \citep{10.1145/2684822.2685324}, which evaluates topic coherence based on the co-occurrence of the topic's top words in a reference corpus. Following standard evaluation protocol, we use Wikipedia as the reference corpus and consider the top 10 words of each topic, with implementation done using the Palmetto package\footnote{https://github.com/dice-group/Palmetto} \citep{10.1145/2684822.2685324}. We report the average $C_V$ score of all learned topics as the overall topic coherence performance. For documents' topical representation (i.e., topic proportion) evaluation, a common practice is to compare the document clusters formed by topic proportions with those formed by the documents' true labels, known as topic alignment. Following previous works \citep{10.5555/3042817.3043005,pham-etal-2024-topicgpt}, we assign each test document to a cluster based on the top-weighted topic of its topical representation, and compute Purity and Normalized Mutual Information (NMI) based on the documents' cluster assignments and their true labels. As Purity and NMI are often reported together and within the same range, we report the average score of both metrics as PN. For all evaluations, we use the model state at the end of the training iteration to compute all evaluation metrics.

\section{More Results}\label{appendix:all_ks}
\setcounter{table}{0} 
\renewcommand{\thetable}{E\arabic{table}}

\subsection{Topic Coherence at Different K}
We illustrate the topic coherence performance ($C_V$) of topic models across different numbers of topics (i.e., $K$) in Table \ref{tab:tc_k}. Overall, using LLM-ITL significantly boosts topic coherence across all cases in our experiments, with minimum gains of +10.4\% (NSTM as the base model on DBpedia at $K=50$) and maximum gains of +86.3\% (ECRTM as the base model on 20News at $K=25$) over the base model.

\subsection{Topic Alignment at Different K}
We illustrate the topic alignment performance (PN) of topic models across different numbers of topics (i.e., $K$) in Table \ref{tab:pn_k}. Overall, using LLM-ITL inherits the topic alignment performance of its base model, with slight changes ranging from -5.2\% (NSTM as the base model on 20News at $K=100$) to +5.2\% (PLDA as the base model on 20News at $K=75$).

\subsection{Topic Diversity (TD)} \label{appendix:TD}
We evaluate the performance of topic diversity (TD) defined by \citet{dieng-etal-2020-topic}, which evaluates the distinctiveness of the top-words within the learned topics. The results across different numbers of topics (i.e., $K$) is illustrated in Table \ref{tab:td_k}. Overall, applying LLM-ITL results in changes to TD ranging from -34.6\% (ECRTM as the base model on 20News at $K=100$) to +55.5\% (NSTM as the base model on 20News at $K=100$) compared to its base model. A further analysis of overall topic quality, which considers both topic coherence and diversity, is described in the following section.

\subsection{Overall Topic Quality (TQ)} \label{appendix:TQ}
\citet{wang2022representing} define topic quality (TQ) as $\text{TQ}:=\text{TC}\times\text{TD}$, serving as an overall indicator of topic quality that  considers both topic coherence and diversity. We illustrate the topic quality performance of topic models across different numbers of topics (i.e., $K$) in Table \ref{tab:tq_k}. Overall, LLM-ITL boosts overall topic quality in all cases in our experiments, with a minimum improvement of 3.7\% (SCHOLAR as the base model on R8 at $K=100$) and a maximum improvement of 90.9\% (CLNTM as the base model on DBpedia at $K=100$) compared to its base mod

\subsection{Purity \& NMI} \label{purity_nmi}
As we report PN (the mean of Purity and NMI) as an overall topic alignment metric in previous sections, here we present the detailed performance of Purity and NMI in Table \ref{tab:tp_k} and Table \ref{tab:tn_k}, respectively. 

\newpage
\onecolumn

\begin{table*}[!t]
    \centering
    \resizebox{\textwidth}{!}{
    \begin{tabular}{lrrr|rrr|rrr|rrr}
    \toprule
    \multirow{3}{*}{\textbf{Model}} & \multicolumn{3}{c}{\textbf{20News}} & \multicolumn{3}{c}{\textbf{R8}} & \multicolumn{3}{c}{\textbf{DBpedia}} & \multicolumn{3}{c}{\textbf{AGNews}}\\
    \cmidrule(l){2-4} \cmidrule(l){5-7} \cmidrule(l){8-10} \cmidrule(l){11-13} 
    & \textbf{K=25} & \textbf{K=75} & \textbf{K=100} & \textbf{K=25} & \textbf{K=75} & \textbf{K=100} & \textbf{K=50} & \textbf{K=75} & \textbf{K=100} &  \textbf{K=50} & \textbf{K=75} & \textbf{K=100}\\
    \midrule
    LDA  & 0.558\small{ ± 0.008} & 0.525\small{ ± 0.005} & 0.510\small{ ± 0.004} & 0.387\small{ ± 0.011} & 0.373\small{ ± 0.003} & 0.374\small{ ± 0.006} & 0.529\small{ ± 0.005} & 0.520\small{ ± 0.010} & 0.506\small{ ± 0.004} & 0.526\small{ ± 0.010} & 0.502\small{ ± 0.003} & 0.480\small{ ± 0.003}\\
    BERTopic & 0.435\small{ ± 0.029} & 0.403\small{ ± 0.012} & 0.390\small{ ± 0.005} & 0.349\small{ ± 0.013} & 0.329\small{ ± 0.008} & 0.333\small{ ± 0.007} & 0.559\small{ ± 0.011} & 0.537\small{ ± 0.007} & 0.516\small{ ± 0.006} & 0.500\small{ ± 0.016} & 0.485\small{ ± 0.004} & 0.479\small{ ± 0.004}\\
    TopicGPT  & NA & NA & NA & NA & NA & NA & NA & NA & NA & NA & NA & NA \\
    \cmidrule(l){1-13}
    NVDM  & 0.299\small{ ± 0.012} & 0.241\small{ ± 0.004} & 0.232\small{ ± 0.004} & 0.324\small{ ± 0.011} & 0.283\small{ ± 0.002} & 0.270\small{ ± 0.002} & 0.274\small{ ± 0.003} & 0.260\small{ ± 0.004} & 0.259\small{ ± 0.005} & 0.289\small{ ± 0.004} & 0.266\small{ ± 0.005} & 0.256\small{ ± 0.007} \\
    + LLM-ITL & 0.439\small{ ± 0.019} & 0.298\small{ ± 0.009} & 0.292\small{ ± 0.004} & 0.438\small{ ± 0.017} & 0.358\small{ ± 0.011} & 0.340\small{ ± 0.006} & 0.365\small{ ± 0.008} & 0.353\small{ ± 0.008} & 0.346\small{ ± 0.006} & 0.357\small{ ± 0.018} & 0.326\small{ ± 0.005} & 0.325\small{ ± 0.005}\\
    & \bm{$\uparrow$} \textbf{46.8\%} & \bm{$\uparrow$} \textbf{23.7\%} & \bm{$\uparrow$} \textbf{25.9\%} & \bm{$\uparrow$} \textbf{35.2\%} & \bm{$\uparrow$} \textbf{26.5\%} & \bm{$\uparrow$} \textbf{25.9\%} & \bm{$\uparrow$} \textbf{33.2\%} & \bm{$\uparrow$} \textbf{35.8\%} & \bm{$\uparrow$} \textbf{33.6\%} & \bm{$\uparrow$} \textbf{23.5\%} & \bm{$\uparrow$} \textbf{22.6\%} & \bm{$\uparrow$} \textbf{27.0\%} \\
     \cmidrule(l){1-13}
    PLDA  & 0.418\small{ ± 0.020} & 0.352\small{ ± 0.010} & 0.336\small{ ± 0.002} & 0.391\small{ ± 0.008} & 0.365\small{ ± 0.006} & 0.348\small{ ± 0.007} & 0.484\small{ ± 0.008} & 0.457\small{ ± 0.009} & 0.438\small{ ± 0.004} & 0.515\small{ ± 0.007} & 0.490\small{ ± 0.012} & 0.470\small{ ± 0.003} \\
    + LLM-ITL & 0.557\small{ ± 0.026} & 0.534\small{ ± 0.010} & 0.506\small{ ± 0.013} & 0.477\small{ ± 0.036} & 0.456\small{ ± 0.011} & 0.452\small{ ± 0.004} & 0.598\small{ ± 0.009} & 0.570\small{ ± 0.021} & 0.568\small{ ± 0.015} & 0.595\small{ ± 0.012} & 0.582\small{ ± 0.012} & 0.581\small{ ± 0.008}\\
    & \bm{$\uparrow$} \textbf{33.3\%} & \bm{$\uparrow$} \textbf{51.7\%} & \bm{$\uparrow$} \textbf{50.6\%} & \bm{$\uparrow$} \textbf{22.0\%} & \bm{$\uparrow$} \textbf{24.9\%} & \bm{$\uparrow$} \textbf{29.9\%} & \bm{$\uparrow$} \textbf{23.6\%} & \bm{$\uparrow$} \textbf{24.7\%} & \bm{$\uparrow$} \textbf{29.7\%} & \bm{$\uparrow$} \textbf{15.5\%} & \bm{$\uparrow$} \textbf{18.8\%} & \bm{$\uparrow$} \textbf{23.6\%} \\
    \cmidrule(l){1-13}
    SCHOLAR & 0.520\small{ ± 0.018} & 0.478\small{ ± 0.009} & 0.472\small{ ± 0.014} & 0.399\small{ ± 0.010} & 0.385\small{ ± 0.005} & 0.376\small{ ± 0.004} & 0.434\small{ ± 0.007} & 0.422\small{ ± 0.015} & 0.403\small{ ± 0.021} & 0.577\small{ ± 0.025} & 0.581\small{ ± 0.010} & 0.541\small{ ± 0.014}\\
    + LLM-ITL & 0.622\small{ ± 0.028} & 0.584\small{ ± 0.008} & 0.562\small{ ± 0.011} & 0.462\small{ ± 0.010} & 0.437\small{ ± 0.006} & 0.424\small{ ± 0.009} & 0.564\small{ ± 0.037} & 0.565\small{ ± 0.024} & 0.556\small{ ± 0.020} & 0.643\small{ ± 0.015} & 0.647\small{ ± 0.005} & 0.627\small{ ± 0.014} \\
    & \bm{$\uparrow$} \textbf{19.6\%} & \bm{$\uparrow$} \textbf{22.2\%} & \bm{$\uparrow$} \textbf{19.1\%} & \bm{$\uparrow$} \textbf{15.8\%} & \bm{$\uparrow$} \textbf{13.5\%} & \bm{$\uparrow$} \textbf{12.8\%} & \bm{$\uparrow$} \textbf{30.0\%} & \bm{$\uparrow$} \textbf{33.9\%} & \bm{$\uparrow$} \textbf{38.0\%} & \bm{$\uparrow$} \textbf{11.4\%} & \bm{$\uparrow$} \textbf{11.4\%} & \bm{$\uparrow$} \textbf{15.9\%}\\
    \cmidrule(l){1-13}
    ETM  & 0.511\small{ ± 0.007} & 0.466\small{ ± 0.005} & 0.461\small{ ± 0.008} & 0.447\small{ ± 0.017} & 0.432\small{ ± 0.011} & 0.431\small{ ± 0.009} & 0.507\small{ ± 0.007} & 0.501\small{ ± 0.006} & 0.496\small{ ± 0.007} & 0.497\small{ ± 0.010} & 0.497\small{ ± 0.004} & 0.480\small{ ± 0.004} \\
    + LLM-ITL & 0.627\small{ ± 0.019} & 0.566\small{ ± 0.007} & 0.553\small{ ± 0.014} & 0.570\small{ ± 0.014} & 0.565\small{ ± 0.010} & 0.580\small{ ± 0.004} & 0.657\small{ ± 0.031} & 0.631\small{ ± 0.014} & 0.626\small{ ± 0.009} & 0.626\small{ ± 0.018} & 0.620\small{ ± 0.012} & 0.618\small{ ± 0.006} \\
    & \bm{$\uparrow$} \textbf{22.7\%} & \bm{$\uparrow$} \textbf{21.5\%} & \bm{$\uparrow$} \textbf{20.0\%} & \bm{$\uparrow$} \textbf{27.5\%} & \bm{$\uparrow$} \textbf{30.8\%} & \bm{$\uparrow$} \textbf{34.6\%} & \bm{$\uparrow$} \textbf{29.6\%} & \bm{$\uparrow$} \textbf{25.9\%} & \bm{$\uparrow$} \textbf{26.2\%} & \bm{$\uparrow$} \textbf{26.0\%} & \bm{$\uparrow$} \textbf{24.7\%} & \bm{$\uparrow$} \textbf{28.8\%}\\
    \cmidrule(l){1-13}
    NSTM & 0.450\small{ ± 0.008} & 0.425\small{ ± 0.003} & 0.424\small{ ± 0.008} & 0.415\small{ ± 0.014} & 0.407\small{ ± 0.006} & 0.402\small{ ± 0.013} & 0.636\small{ ± 0.008} & 0.621\small{ ± 0.014} & 0.616\small{ ± 0.006} & 0.596\small{ ± 0.015} & 0.583\small{ ± 0.015} & 0.575\small{ ± 0.018}\\
    + LLM-ITL & 0.510\small{ ± 0.027} & 0.524\small{ ± 0.008} & 0.518\small{ ± 0.011} & 0.549\small{ ± 0.011} & 0.544\small{ ± 0.021} & 0.534\small{ ± 0.016} & 0.702\small{ ± 0.009} & 0.688\small{ ± 0.009} & 0.690\small{ ± 0.014} & 0.661\small{ ± 0.021} & 0.670\small{ ± 0.008} & 0.649\small{ ± 0.012} \\
    & \bm{$\uparrow$} \textbf{13.3\%} & \bm{$\uparrow$} \textbf{23.3\%} & \bm{$\uparrow$} \textbf{22.2\%} & \bm{$\uparrow$} \textbf{32.3\%} & \bm{$\uparrow$} \textbf{33.7\%} & \bm{$\uparrow$} \textbf{32.8\%} & \bm{$\uparrow$} \textbf{10.4\%} & \bm{$\uparrow$} \textbf{10.8\%} & \bm{$\uparrow$} \textbf{12.0\%} & \bm{$\uparrow$} \textbf{10.9\%} & \bm{$\uparrow$} \textbf{14.9\%} & \bm{$\uparrow$} \textbf{12.9\%} \\
    \cmidrule(l){1-13}
    CLNTM  & 0.501\small{ ± 0.010} & 0.478\small{ ± 0.007} & 0.469\small{ ± 0.010} & 0.387\small{ ± 0.005} & 0.346\small{ ± 0.004} & 0.354\small{ ± 0.002} & 0.414\small{ ± 0.030} & 0.394\small{ ± 0.014} & 0.393\small{ ± 0.011} & 0.549\small{ ± 0.010} & 0.567\small{ ± 0.012} & 0.569\small{ ± 0.010}\\
    + LLM-ITL & 0.623\small{ ± 0.011} & 0.587\small{ ± 0.011} & 0.580\small{ ± 0.006} & 0.446\small{ ± 0.006} & 0.422\small{ ± 0.008} & 0.411\small{ ± 0.004} & 0.556\small{ ± 0.004} & 0.540\small{ ± 0.020} & 0.547\small{ ± 0.008} & 0.613\small{ ± 0.003} & 0.628\small{ ± 0.005} & 0.658\small{ ± 0.013} \\
    & \bm{$\uparrow$} \textbf{24.4\%} & \bm{$\uparrow$} \textbf{22.8\%} & \bm{$\uparrow$} \textbf{23.7\%} & \bm{$\uparrow$} \textbf{15.2\%} & \bm{$\uparrow$} \textbf{22.0\%} & \bm{$\uparrow$} \textbf{16.1\%} & \bm{$\uparrow$} \textbf{34.3\%} & \bm{$\uparrow$} \textbf{37.1\%} & \bm{$\uparrow$} \textbf{39.2\%} & \bm{$\uparrow$} \textbf{11.7\%} & \bm{$\uparrow$} \textbf{10.8\%} & \bm{$\uparrow$} \textbf{15.6\%} \\
    \cmidrule(l){1-13}
    WeTe & 0.488\small{ ± 0.010} & 0.505\small{ ± 0.017} & 0.498\small{ ± 0.013} & 0.534\small{ ± 0.008} & 0.534\small{ ± 0.011} & 0.533\small{ ± 0.011} & 0.557\small{ ± 0.012} & 0.526\small{ ± 0.005} & 0.516\small{ ± 0.015} & 0.587\small{ ± 0.007} & 0.562\small{ ± 0.003} & 0.556\small{ ± 0.005} \\
   + LLM-ITL & 0.539\small{ ± 0.030} & 0.578\small{ ± 0.022} & 0.556\small{ ± 0.021} & 0.604\small{ ± 0.015} & 0.605\small{ ± 0.012} & 0.601\small{ ± 0.008} & 0.650\small{ ± 0.009} & 0.660\small{ ± 0.011} & 0.665\small{ ± 0.018} & 0.655\small{ ± 0.007} & 0.689\small{ ± 0.010} & 0.693\small{ ± 0.015} \\
   & \bm{$\uparrow$} \textbf{10.5\%} & \bm{$\uparrow$} \textbf{14.5\%} & \bm{$\uparrow$} \textbf{11.6\%} & \bm{$\uparrow$} \textbf{13.1\%} & \bm{$\uparrow$} \textbf{13.3\%} & \bm{$\uparrow$} \textbf{12.8\%} & \bm{$\uparrow$} \textbf{16.7\%} & \bm{$\uparrow$} \textbf{25.5\%} & \bm{$\uparrow$} \textbf{28.9\%} & \bm{$\uparrow$} \textbf{11.6\%} & \bm{$\uparrow$} \textbf{22.6\%} & \bm{$\uparrow$} \textbf{24.6\%} \\
   \cmidrule(l){1-13}
   ECRTM & 0.350\small{ ± 0.013} & 0.324\small{ ± 0.007} & 0.313\small{ ± 0.003} & 0.279\small{ ± 0.015} & 0.287\small{ ± 0.013} & 0.288\small{ ± 0.009} & 0.520\small{ ± 0.016} & 0.436\small{ ± 0.014} & 0.407\small{ ± 0.008} & 0.408\small{ ± 0.026} & 0.382\small{ ± 0.011} & 0.352\small{ ± 0.014}\\
   + LLM-ITL & 0.652\small{ ± 0.026} & 0.518\small{ ± 0.017} & 0.520\small{ ± 0.023} & 0.352\small{ ± 0.012} & 0.354\small{ ± 0.008} & 0.362\small{ ± 0.008} & 0.630\small{ ± 0.006} & 0.563\small{ ± 0.010} & 0.521\small{ ± 0.010} & 0.475\small{ ± 0.019} & 0.451\small{ ± 0.021} & 0.426\small{ ± 0.012}\\
   & \bm{$\uparrow$} \textbf{86.3\%} & \bm{$\uparrow$} \textbf{59.9\%} & \bm{$\uparrow$} \textbf{66.1\%} & \bm{$\uparrow$} \textbf{26.2\%} & \bm{$\uparrow$} \textbf{23.3\%} & \bm{$\uparrow$} \textbf{25.7\%} & \bm{$\uparrow$} \textbf{21.2\%} & \bm{$\uparrow$} \textbf{29.1\%} & \bm{$\uparrow$} \textbf{28.0\%} & \bm{$\uparrow$} \textbf{16.4\%} & \bm{$\uparrow$} \textbf{18.1\%} & \bm{$\uparrow$} \textbf{21.0\%}\\
    \bottomrule
    \end{tabular}
    }
    \caption{Topic coherence ($C_V$) at other settings of $K$.} \label{tab:tc_k}
\end{table*}

\hfill

\begin{table*}[!h]
    \centering
    \resizebox{\textwidth}{!}{
    \begin{tabular}{lrrr|rrr|rrr|rrr}
    \toprule
    \multirow{3}{*}{\textbf{Model}} & \multicolumn{3}{c}{\textbf{20News}} & \multicolumn{3}{c}{\textbf{R8}} & \multicolumn{3}{c}{\textbf{DBpedia}} & \multicolumn{3}{c}{\textbf{AGNews}}\\
    \cmidrule(l){2-4} \cmidrule(l){5-7} \cmidrule(l){8-10} \cmidrule(l){11-13} 
    & \textbf{K=25} & \textbf{K=75} & \textbf{K=100} & \textbf{K=25} & \textbf{K=75} & \textbf{K=100} & \textbf{K=50} & \textbf{K=75} & \textbf{K=100} & \textbf{K=50} & \textbf{K=75} & \textbf{K=100}\\
    \midrule
    LDA  & 0.460\small{ ± 0.011} & 0.488\small{ ± 0.006} & 0.489\small{ ± 0.012} & 0.714\small{ ± 0.008} & 0.699\small{ ± 0.002} & 0.689\small{ ± 0.006} & 0.748\small{ ± 0.007} & 0.738\small{ ± 0.005} & 0.730\small{ ± 0.007} & 0.576\small{ ± 0.005} & 0.570\small{ ± 0.004} & 0.562\small{ ± 0.005}\\
    BERTopic  & 0.289\small{ ± 0.021} & 0.372\small{ ± 0.004} & 0.390\small{ ± 0.004} & 0.732\small{ ± 0.011} & 0.689\small{ ± 0.006} & 0.685\small{ ± 0.006} & 0.739\small{ ± 0.002} & 0.716\small{ ± 0.010} & 0.695\small{ ± 0.003} & 0.472\small{ ± 0.004} & 0.484\small{ ± 0.010} & 0.495\small{ ± 0.004}\\
    TopicGPT  & 0.363\small{ ± 0.000} & 0.363\small{ ± 0.000} & 0.363\small{ ± 0.000} & 0.410\small{ ± 0.000} & 0.410\small{ ± 0.000} & 0.410\small{ ± 0.000} & 0.706\small{ ± 0.000} & 0.706\small{ ± 0.000} & 0.706\small{ ± 0.000} & 0.634\small{ ± 0.000} & 0.634\small{ ± 0.000} & 0.634\small{ ± 0.000}\\
    \cmidrule(l){1-13}
    NVDM  & 0.146\small{ ± 0.005} & 0.145\small{ ± 0.003} & 0.153\small{ ± 0.003} & 0.380\small{ ± 0.007} & 0.353\small{ ± 0.013} & 0.359\small{ ± 0.003} & 0.150\small{ ± 0.004} & 0.151\small{ ± 0.003} & 0.156\small{ ± 0.004} & 0.223\small{ ± 0.004} & 0.212\small{ ± 0.004} & 0.211\small{ ± 0.004} \\
    + LLM-ITL & 0.143\small{ ± 0.005} & 0.142\small{ ± 0.005} & 0.153\small{ ± 0.004} & 0.377\small{ ± 0.008} & 0.355\small{ ± 0.006} & 0.358\small{ ± 0.003} & 0.150\small{ ± 0.004} & 0.150\small{ ± 0.003} & 0.152\small{ ± 0.003} & 0.222\small{ ± 0.001} & 0.212\small{ ± 0.003} & 0.211\small{ ± 0.004} \\
    & $\downarrow$ 2.1\% & $\downarrow$ 2.1\% & \bm{$\uparrow$} \textbf{0.0\%} & $\downarrow$ 0.8\% & \bm{$\uparrow$} \textbf{0.6\%} & $\downarrow$ 0.3\% & \bm{$\uparrow$} \textbf{0.0\%} & $\downarrow$ 0.7\% & $\downarrow$ 2.6\% & $\downarrow$ 0.4\% & \bm{$\uparrow$} \textbf{0.0\%} & \bm{$\uparrow$} \textbf{0.0\%} \\
     \cmidrule(l){1-13}
    PLDA  & 0.231\small{ ± 0.003} & 0.290\small{ ± 0.005} & 0.303\small{ ± 0.004} & 0.523\small{ ± 0.014} & 0.520\small{ ± 0.007} & 0.520\small{ ± 0.006} & 0.643\small{ ± 0.004} & 0.620\small{ ± 0.008} & 0.610\small{ ± 0.005} & 0.474\small{ ± 0.005} & 0.475\small{ ± 0.004} & 0.471\small{ ± 0.005} \\
    + LLM-ITL & 0.241\small{ ± 0.003} & 0.305\small{ ± 0.010} & 0.313\small{ ± 0.011} & 0.533\small{ ± 0.009} & 0.518\small{ ± 0.008} & 0.514\small{ ± 0.006} & 0.641\small{ ± 0.010} & 0.624\small{ ± 0.006} & 0.611\small{ ± 0.005} & 0.472\small{ ± 0.004} & 0.470\small{ ± 0.003} & 0.466\small{ ± 0.001}\\
    & \bm{$\uparrow$} \textbf{4.3\%} & \bm{$\uparrow$} \textbf{5.2\%} & \bm{$\uparrow$} \textbf{3.3\%} & \bm{$\uparrow$} \textbf{1.9\%} & $\downarrow$ 0.4\% & $\downarrow$ 1.2\% & $\downarrow$ 0.3\% & \bm{$\uparrow$} \textbf{0.6\%} & \bm{$\uparrow$} \textbf{0.2\%} & $\downarrow$ 0.4\% & $\downarrow$ 1.1\% & $\downarrow$ 1.1\% \\
    \cmidrule(l){1-13}
    SCHOLAR &  0.572\small{ ± 0.010} & 0.580\small{ ± 0.008} & 0.588\small{ ± 0.004} & 0.717\small{ ± 0.017} & 0.686\small{ ± 0.006} & 0.681\small{ ± 0.005} & 0.827\small{ ± 0.016} & 0.794\small{ ± 0.010} & 0.680\small{ ± 0.114} & 0.587\small{ ± 0.049} & 0.426\small{ ± 0.048} & 0.389\small{ ± 0.067} \\
    + LLM-ITL & 0.562\small{ ± 0.011} & 0.578\small{ ± 0.006} & 0.587\small{ ± 0.007} & 0.716\small{ ± 0.018} & 0.682\small{ ± 0.006} & 0.675\small{ ± 0.004} & 0.828\small{ ± 0.014} & 0.773\small{ ± 0.015} & 0.701\small{ ± 0.055} & 0.567\small{ ± 0.053} & 0.410\small{ ± 0.073} & 0.382\small{ ± 0.076} \\
    & $\downarrow$ 1.7\% & $\downarrow$ 0.3\% & $\downarrow$ 0.2\% & $\downarrow$ 0.1\% & $\downarrow$ 0.6\% & $\downarrow$ 0.9\% & \bm{$\uparrow$} \textbf{0.1\%} & $\downarrow$ 2.6\% & \bm{$\uparrow$} \textbf{3.1\%} & $\downarrow$ 3.4\% & $\downarrow$ 3.8\% & $\downarrow$ 1.8\% \\
    \cmidrule(l){1-13}
    ETM  & 0.363\small{ ± 0.007} & 0.434\small{ ± 0.008} & 0.456\small{ ± 0.013} & 0.679\small{ ± 0.018} & 0.666\small{ ± 0.007} & 0.664\small{ ± 0.005} & 0.751\small{ ± 0.010} & 0.747\small{ ± 0.005} & 0.739\small{ ± 0.006} & 0.553\small{ ± 0.003} & 0.555\small{ ± 0.006} & 0.549\small{ ± 0.003} \\
    + LLM-ITL & 0.352\small{ ± 0.009} & 0.426\small{ ± 0.012} & 0.441\small{ ± 0.012} & 0.690\small{ ± 0.011} & 0.673\small{ ± 0.011} & 0.666\small{ ± 0.019} & 0.740\small{ ± 0.005} & 0.738\small{ ± 0.014} & 0.744\small{ ± 0.008} & 0.558\small{ ± 0.005} & 0.560\small{ ± 0.008} & 0.553\small{ ± 0.008} \\
    & $\downarrow$ 3.0\% & $\downarrow$ 1.8\% & $\downarrow$ 3.3\% & \bm{$\uparrow$} \textbf{1.6\%} & \bm{$\uparrow$} \textbf{1.1\%} & \bm{$\uparrow$} \textbf{0.3\%} & $\downarrow$ 1.5\% & $\downarrow$ 1.2\% & \bm{$\uparrow$} \textbf{0.7\%} & \bm{$\uparrow$} \textbf{0.9\%} & \bm{$\uparrow$} \textbf{0.9\%} & \bm{$\uparrow$} \textbf{0.7\%} \\
    \cmidrule(l){1-13}
    NSTM & 0.365\small{ ± 0.013} & 0.382\small{ ± 0.006} & 0.405\small{ ± 0.007} & 0.674\small{ ± 0.004} & 0.662\small{ ± 0.004} & 0.662\small{ ± 0.007} & 0.739\small{ ± 0.007} & 0.739\small{ ± 0.003} & 0.738\small{ ± 0.004} & 0.576\small{ ± 0.007} & 0.570\small{ ± 0.006} & 0.569\small{ ± 0.005} \\
    + LLM-ITL & 0.351\small{ ± 0.012} & 0.366\small{ ± 0.008} & 0.384\small{ ± 0.009} & 0.681\small{ ± 0.012} & 0.669\small{ ± 0.003} & 0.669\small{ ± 0.009} & 0.741\small{ ± 0.009} & 0.738\small{ ± 0.006} & 0.736\small{ ± 0.002} & 0.576\small{ ± 0.006} & 0.566\small{ ± 0.006} & 0.566\small{ ± 0.004} \\
    & $\downarrow$ 3.8\% & $\downarrow$ 4.2\% & $\downarrow$ 5.2\% & \bm{$\uparrow$} \textbf{1.0\%} & \bm{$\uparrow$} \textbf{1.1\%} & \bm{$\uparrow$} \textbf{1.1\%} & \bm{$\uparrow$} \textbf{0.3\%} & $\downarrow$ 0.1\% & $\downarrow$ 0.3\% & \bm{$\uparrow$} \textbf{0.0\%} & $\downarrow$ 0.7\% & $\downarrow$ 0.5\% \\
    \cmidrule(l){1-13}
    CLNTM  & 0.536\small{ ± 0.007} & 0.581\small{ ± 0.008} & 0.586\small{ ± 0.006} & 0.691\small{ ± 0.010} & 0.680\small{ ± 0.009} & 0.680\small{ ± 0.006} & 0.542\small{ ± 0.039} & 0.438\small{ ± 0.019} & 0.427\small{ ± 0.062} & 0.578\small{ ± 0.047} & 0.487\small{ ± 0.055} & 0.438\small{ ± 0.035} \\
    + LLM-ITL & 0.529\small{ ± 0.013} & 0.585\small{ ± 0.005} & 0.585\small{ ± 0.007} & 0.690\small{ ± 0.009} & 0.680\small{ ± 0.008} & 0.676\small{ ± 0.006} & 0.532\small{ ± 0.032} & 0.438\small{ ± 0.053} & 0.428\small{ ± 0.034} & 0.579\small{ ± 0.026} & 0.468\small{ ± 0.073} & 0.453\small{ ± 0.104} \\
    & $\downarrow$ 1.3\% & \bm{$\uparrow$} \textbf{0.7\%} & $\downarrow$ 0.2\% & $\downarrow$ 0.1\% & \bm{$\uparrow$} \textbf{0.0\%} & $\downarrow$ 0.6\% & $\downarrow$ 1.8\% & \bm{$\uparrow$} \textbf{0.0\%} & \bm{$\uparrow$} \textbf{0.2\%} & \bm{$\uparrow$} \textbf{0.2\%} & $\downarrow$ 3.9\% & \bm{$\uparrow$} \textbf{3.4\%}  \\
    \cmidrule(l){1-13}
    WeTe  & 0.284\small{ ± 0.005} & 0.329\small{ ± 0.007} & 0.339\small{ ± 0.007} & 0.678\small{ ± 0.002} & 0.670\small{ ± 0.007} & 0.649\small{ ± 0.002} & 0.724\small{ ± 0.009} & 0.726\small{ ± 0.010} & 0.725\small{ ± 0.010} & 0.502\small{ ± 0.006} & 0.502\small{ ± 0.003} & 0.487\small{ ± 0.005} \\
    + LLM-ITL & 0.278\small{ ± 0.003} & 0.335\small{ ± 0.011} & 0.336\small{ ± 0.006} & 0.685\small{ ± 0.003} & 0.678\small{ ± 0.009} & 0.653\small{ ± 0.005} & 0.722\small{ ± 0.014} & 0.727\small{ ± 0.008} & 0.727\small{ ± 0.003} & 0.501\small{ ± 0.005} & 0.489\small{ ± 0.005} & 0.500\small{ ± 0.004}\\
    & $\downarrow$ 2.1\% & \bm{$\uparrow$} \textbf{1.8\%} & $\downarrow$ 0.9\% & \bm{$\uparrow$} \textbf{1.0\%} & \bm{$\uparrow$} \textbf{1.2\%} & \bm{$\uparrow$} \textbf{0.6\%} & $\downarrow$ 0.3\% & \bm{$\uparrow$} \textbf{0.1\%} & \bm{$\uparrow$} \textbf{0.3\%} & $\downarrow$ 0.2\% & $\downarrow$ 2.6\% & \bm{$\uparrow$} \textbf{2.7\%}  \\
    \cmidrule(l){1-13}
    ECRTM  & 0.520\small{ ± 0.008} & 0.533\small{ ± 0.004} & 0.535\small{ ± 0.003} & 0.652\small{ ± 0.015} & 0.670\small{ ± 0.009} & 0.665\small{ ± 0.008} & 0.750\small{ ± 0.007} & 0.745\small{ ± 0.013} & 0.772\small{ ± 0.006} & 0.543\small{ ± 0.010} & 0.536\small{ ± 0.009} & 0.539\small{ ± 0.007}\\
    + LLM-ITL & 0.519\small{ ± 0.009} & 0.527\small{ ± 0.005} & 0.532\small{ ± 0.003} & 0.644\small{ ± 0.015} & 0.666\small{ ± 0.013} & 0.666\small{ ± 0.009} & 0.758\small{ ± 0.008} & 0.776\small{ ± 0.013} & 0.776\small{ ± 0.019} & 0.541\small{ ± 0.008} & 0.539\small{ ± 0.012} & 0.524\small{ ± 0.061} \\
    &  $\downarrow$ 0.2\% & $\downarrow$ 1.1\% & $\downarrow$ 0.6\% & $\downarrow$ 1.2\% & $\downarrow$ 0.6\% & \bm{$\uparrow$} \textbf{0.2\%} & \bm{$\uparrow$} \textbf{1.1\%} & \bm{$\uparrow$} \textbf{4.2\%} & \bm{$\uparrow$} \textbf{0.5\%} & $\downarrow$ 0.4\% & \bm{$\uparrow$} \textbf{0.6\%} & $\downarrow$ 2.8\% \\
    \bottomrule
    \end{tabular}
    }
    \caption{Topic alignment (PN) at other settings of $K$.} \label{tab:pn_k}
\end{table*}

\newpage

\begin{table*}[!t]
    \centering
    \resizebox{\textwidth}{!}{
    \begin{tabular}{lcccc|cccc|cccc|cccc}
    \toprule
    \multirow{2}{*}{\textbf{Model}} & \multicolumn{4}{c}{\textbf{20News}} & \multicolumn{4}{c}{\textbf{R8}} & \multicolumn{4}{c}{\textbf{DBpedia}} & \multicolumn{4}{c}{\textbf{AGNews}}\\
    \cmidrule(l){2-5} \cmidrule(l){6-9} \cmidrule(l){10-13} \cmidrule(l){14-17} 
    & \textbf{25} & \textbf{50} & \textbf{75} & \textbf{100} & \textbf{25} & \textbf{50} & \textbf{75} & \textbf{100} &  \textbf{25} & \textbf{50} & \textbf{75} & \textbf{100} & \textbf{25} & \textbf{50} & \textbf{75} & \textbf{100}\\
    \midrule
    LDA  & 0.854 & 0.752 & 0.701 & 0.667 & 0.413 & 0.294 & 0.244 & 0.223 & 0.726 & 0.612 & 0.555 & 0.527 & 0.606 & 0.446 & 0.366 & 0.331 \\
    BERTopic  & 0.971 & 0.951 & 0.934 & 0.905 & 0.735 & 0.620 & 0.537 & 0.507 & 0.958 & 0.936 & 0.899 & 0.859 & 0.949 & 0.934 & 0.925 & 0.905 \\
    TopicGPT  & - & - & - & - & - & - & - & - & - & - & - & - & - & - & - & -\\
    \cmidrule(l){1-17}
    NVDM  & 0.993 & 0.985 & 0.956 & 0.931 & 0.802 & 0.801 & 0.789 & 0.790 & 0.982 & 0.964 & 0.919 & 0.875 & 0.962 & 0.940 & 0.896 & 0.847 \\
    + LLM-ITL & 0.970 & 0.966 & 0.930 & 0.896 & 0.834 & 0.811 & 0.765 & 0.714 & 0.981 & 0.940 & 0.878 & 0.829 & 0.962 & 0.929 & 0.871 & 0.813\\
    & $\downarrow$ 2.3\% & $\downarrow$ 1.9\% & $\downarrow$ 2.7\% & $\downarrow$ 3.8\% & \bm{$\uparrow$} \textbf{4.0\%} & \bm{$\uparrow$} \textbf{1.2\%} & $\downarrow$ 3.0\% & $\downarrow$ 9.6\% & $\downarrow$ 0.1\% & $\downarrow$ 2.5\% & $\downarrow$ 4.5\% & $\downarrow$ 5.3\% & \bm{$\uparrow$} \textbf{0.0\%} & $\downarrow$ 1.2\% & $\downarrow$ 2.8\% & $\downarrow$ 4.0\% \\
     \cmidrule(l){1-17}
    PLDA  & 0.854 & 0.847 & 0.857 & 0.852 & 0.654 & 0.506 & 0.454 & 0.424 & 0.886 & 0.886 & 0.852 & 0.821 & 0.837 & 0.837 & 0.771 & 0.717\\
    + LLM-ITL & 0.870 & 0.816 & 0.763 & 0.767 & 0.626 & 0.508 & 0.463 & 0.422 & 0.812 & 0.812 & 0.752 & 0.713 & 0.796 & 0.796 & 0.730 & 0.689\\
    & \bm{$\uparrow$} \textbf{1.9\%} & $\downarrow$ 3.7\% & $\downarrow$ 11.0\% & $\downarrow$ 10.0\% & $\downarrow$ 4.3\% & \bm{$\uparrow$} \textbf{0.4\%} & \bm{$\uparrow$} \textbf{2.0\%} & $\downarrow$ 0.5\% & $\downarrow$ 8.4\% & $\downarrow$ 8.4\% & $\downarrow$ 11.7\% & $\downarrow$ 13.2\% & $\downarrow$ 4.9\% & $\downarrow$ 4.9\% & $\downarrow$ 5.3\% & $\downarrow$ 3.9\% \\
    \cmidrule(l){1-17}
    SCHOLAR & 0.986 & 0.920 & 0.860 & 0.814 & 0.695 & 0.509 & 0.411 & 0.363 & 0.965 & 0.491 & 0.319 & 0.246 & 0.951 & 0.608 & 0.398 & 0.263 \\ 
    + LLM-ITL & 0.958 & 0.863 & 0.793 & 0.736 & 0.662 & 0.475 & 0.385 & 0.332 & 0.926 & 0.520 & 0.392 & 0.330 & 0.903 & 0.605 & 0.458 & 0.292 \\
    & $\downarrow$ 2.8\% & $\downarrow$ 6.2\% & $\downarrow$ 7.8\% & $\downarrow$ 9.6\% & $\downarrow$ 4.7\% & $\downarrow$ 6.7\% & $\downarrow$ 6.3\% & $\downarrow$ 8.5\% & $\downarrow$ 4.0\% & \bm{$\uparrow$} \textbf{5.9\%} & \bm{$\uparrow$} \textbf{22.9\%} & \bm{$\uparrow$} \textbf{34.1\%} & $\downarrow$ 5.0\% & $\downarrow$ 0.5\% & \bm{$\uparrow$} \textbf{15.1\%} & \bm{$\uparrow$} \textbf{11.0\%} \\
    \cmidrule(l){1-17}
    ETM  & 0.890 & 0.786 & 0.687 & 0.604 & 0.688 & 0.521 & 0.444 & 0.373 & 0.907 & 0.864 & 0.762 & 0.644 & 0.932 & 0.828 & 0.738 & 0.668 \\
    + LLM-ITL & 0.940 & 0.832 & 0.754 & 0.695 & 0.676 & 0.518 & 0.441 & 0.377 & 0.974 & 0.909 & 0.837 & 0.770 & 0.909 & 0.733 & 0.636 & 0.578 \\
    & \bm{$\uparrow$} \textbf{5.6\%} & \bm{$\uparrow$} \textbf{5.9\%} & \bm{$\uparrow$} \textbf{9.8\%} & \bm{$\uparrow$} \textbf{15.1\%} & $\downarrow$ 1.7\% & $\downarrow$ 0.6\% & $\downarrow$ 0.7\% & \bm{$\uparrow$} \textbf{1.1\%} & \bm{$\uparrow$} \textbf{7.4\%} & \bm{$\uparrow$} \textbf{5.2\%} & \bm{$\uparrow$} \textbf{9.8\%} & \bm{$\uparrow$} \textbf{19.6\%} & $\downarrow$ 2.5\% & $\downarrow$ 11.5\% & $\downarrow$ 13.8\% & $\downarrow$ 13.5\%  \\
    \cmidrule(l){1-17}
    NSTM & 0.648 & 0.507 & 0.385 & 0.330 & 0.460 & 0.320 & 0.244 & 0.227 & 0.854 & 0.768 & 0.679 & 0.621 & 0.831 & 0.743 & 0.701 & 0.703 \\
    + LLM-ITL & 0.718 & 0.593 & 0.564 & 0.513 & 0.446 & 0.311 & 0.296 & 0.224 & 0.866 & 0.734 & 0.683 & 0.602 & 0.843 & 0.749 & 0.731 & 0.649 \\
    & \bm{$\uparrow$} \textbf{10.8\%} & \bm{$\uparrow$} \textbf{17.0\%} & \bm{$\uparrow$} \textbf{46.5\%} & \bm{$\uparrow$} \textbf{55.5\%} & $\downarrow$ 3.0\% & $\downarrow$ 2.8\% & \bm{$\uparrow$} \textbf{21.3\%} & $\downarrow$ 1.3\% & \bm{$\uparrow$} \textbf{1.4\%} & $\downarrow$ 4.4\% & \bm{$\uparrow$} \textbf{0.6\%} & $\downarrow$ 3.1\% & \bm{$\uparrow$} \textbf{1.4\%} & \bm{$\uparrow$} \textbf{0.8\%} & \bm{$\uparrow$} \textbf{4.3\%} & $\downarrow$ 7.7\% \\
    \cmidrule(l){1-17}
    CLNTM  & 0.978 & 0.908 & 0.850 & 0.790 & 0.654 & 0.448 & 0.355 & 0.315 & 0.851 & 0.452 & 0.293 & 0.224 & 0.889 & 0.404 & 0.272 & 0.220 \\
    + LLM-ITL & 0.936 & 0.846 & 0.779 & 0.718 & 0.699 & 0.489 & 0.386 & 0.345 & 0.842 & 0.491 & 0.358 & 0.308 & 0.820 & 0.427 & 0.300 & 0.274 \\
    & $\downarrow$ 4.3\% & $\downarrow$ 6.8\% & $\downarrow$ 8.4\% & $\downarrow$ 9.1\% & \bm{$\uparrow$} \textbf{6.9\%} & \bm{$\uparrow$} \textbf{9.2\%} & \bm{$\uparrow$} \textbf{8.7\%} & \bm{$\uparrow$} \textbf{9.5\%} & $\downarrow$ 1.1\% & \bm{$\uparrow$} \textbf{8.6\%} & \bm{$\uparrow$} \textbf{22.2\%} & \bm{$\uparrow$} \textbf{37.5\%} & $\downarrow$ 7.8\% & \bm{$\uparrow$} \textbf{5.7\%} & \bm{$\uparrow$} \textbf{10.3\%} & \bm{$\uparrow$} \textbf{24.5\%} \\
    \cmidrule(l){1-17}
    WeTe  & 0.769 & 0.437 & 0.382 & 0.309 & 0.814 & 0.617 & 0.467 & 0.382 & 0.950 & 0.750 & 0.620 & 0.569 & 0.962 & 0.884 & 0.710 & 0.691 \\
    + LLM-ITL & 0.879 & 0.617 & 0.538 & 0.448 & 0.834 & 0.610 & 0.488 & 0.424 & 0.950 & 0.816 & 0.739 & 0.619 & 0.942 & 0.872 & 0.738 & 0.738 \\
    & \bm{$\uparrow$} \textbf{14.3\%} & \bm{$\uparrow$} \textbf{41.2\%} & \bm{$\uparrow$} \textbf{40.8\%} & \bm{$\uparrow$} \textbf{45.0\%} & \bm{$\uparrow$} \textbf{2.5\%} & $\downarrow$ 1.1\% & \bm{$\uparrow$} \textbf{4.5\%} & \bm{$\uparrow$} \textbf{11.0\%} & \bm{$\uparrow$} \textbf{0.0\%} & \bm{$\uparrow$} \textbf{8.8\%} & \bm{$\uparrow$} \textbf{19.2\%} & \bm{$\uparrow$} \textbf{8.8\%} & $\downarrow$ 2.1\% & $\downarrow$ 1.4\% & \bm{$\uparrow$} \textbf{3.9\%} & \bm{$\uparrow$} \textbf{6.8\%} \\
    \cmidrule(l){1-17}
    ECRTM  & 0.981 & 0.986 & 0.962 & 0.965 & 0.979 & 0.970 & 0.969 & 0.976 & 0.966 & 0.934 & 0.973 & 0.986 & 1.000 & 1.000 & 0.999 & 0.999 \\
    + LLM-ITL & 0.769 & 0.747 & 0.701 & 0.631 & 0.954 & 0.958 & 0.950 & 0.932 & 0.926 & 0.898 & 0.926 & 0.933 & 1.000 & 0.976 & 0.954 & 0.928 \\
    &  $\downarrow$ 21.6\% & $\downarrow$ 24.2\% & $\downarrow$ 27.1\% & $\downarrow$ 34.6\% & $\downarrow$ 2.6\% & $\downarrow$ 1.2\% & $\downarrow$ 2.0\% & $\downarrow$ 4.5\% & $\downarrow$ 4.1\% & $\downarrow$ 3.9\% & $\downarrow$ 4.8\% & $\downarrow$ 5.4\% & \bm{$\uparrow$} \textbf{0.0\%} & $\downarrow$ 2.4\% & $\downarrow$ 4.5\% & $\downarrow$ 7.1\% \\
    \bottomrule
    \end{tabular}
    }
    \caption{Topic diversity (TD) at different $K$. Std. values have been omitted to save space in this table.} \label{tab:td_k}
\end{table*}

\hfill

\begin{table*}[!h]
    \centering
    \resizebox{\textwidth}{!}{
    \begin{tabular}{lcccc|cccc|cccc|cccc}
    \toprule
    \multirow{2}{*}{\textbf{Model}} & \multicolumn{4}{c}{\textbf{20News}} & \multicolumn{4}{c}{\textbf{R8}} & \multicolumn{4}{c}{\textbf{DBpedia}} & \multicolumn{4}{c}{\textbf{AGNews}}\\
    \cmidrule(l){2-5} \cmidrule(l){6-9} \cmidrule(l){10-13} \cmidrule(l){14-17} 
    & \textbf{25} & \textbf{50} & \textbf{75} & \textbf{100} & \textbf{25} & \textbf{50} & \textbf{75} & \textbf{100} &  \textbf{25} & \textbf{50} & \textbf{75} & \textbf{100} & \textbf{25} & \textbf{50} & \textbf{75} & \textbf{100}\\
    \midrule
    LDA  & 0.477 & 0.397 & 0.368 & 0.340 & 0.160 & 0.113 & 0.091 & 0.083 & 0.390 & 0.324 & 0.289 & 0.267 & 0.329 & 0.234 & 0.184 & 0.159\\
    BERTopic  & 0.423 & 0.385 & 0.376 & 0.353 & 0.257 & 0.200 & 0.177 & 0.169 & 0.523 & 0.523 & 0.483 & 0.443 & 0.480 & 0.467 & 0.449 & 0.434 \\
    TopicGPT  & - & - & - & - & - & - & - & - & - & - & - & - & - & - & - & -\\
    \cmidrule(l){1-17}
    NVDM  & 0.297 & 0.257 & 0.230 & 0.216 & 0.260 & 0.239 & 0.224 & 0.213 & 0.323 & 0.264 & 0.239 & 0.226 & 0.342 & 0.272 & 0.238 & 0.217 \\
    + LLM-ITL & 0.426 & 0.324 & 0.277 & 0.262 & 0.366 & 0.303 & 0.274 & 0.243 & 0.470 & 0.343 & 0.310 & 0.287 & 0.438 & 0.332 & 0.284 & 0.264 \\
    & \bm{$\uparrow$} \textbf{43.4\%} & \bm{$\uparrow$} \textbf{26.1\%} & \bm{$\uparrow$} \textbf{20.4\%} & \bm{$\uparrow$} \textbf{21.3\%} & \bm{$\uparrow$} \textbf{40.8\%} & \bm{$\uparrow$} \textbf{26.8\%} & \bm{$\uparrow$} \textbf{22.3\%} & \bm{$\uparrow$} \textbf{14.1\%} & \bm{$\uparrow$} \textbf{45.5\%} & \bm{$\uparrow$} \textbf{29.9\%} & \bm{$\uparrow$} \textbf{29.7\%} & \bm{$\uparrow$} \textbf{27.0\%} & \bm{$\uparrow$} \textbf{28.1\%} & \bm{$\uparrow$} \textbf{22.1\%} & \bm{$\uparrow$} \textbf{19.3\%} & \bm{$\uparrow$} \textbf{21.7\%}\\
     \cmidrule(l){1-17}
    PLDA  & 0.357 & 0.312 & 0.302 & 0.287 & 0.256 & 0.190 & 0.166 & 0.148 & 0.479 & 0.429 & 0.390 & 0.360 & 0.522 & 0.431 & 0.378 & 0.337 \\
    + LLM-ITL & 0.484 & 0.428 & 0.407 & 0.388 & 0.300 & 0.237 & 0.211 & 0.191 & 0.575 & 0.485 & 0.429 & 0.405 & 0.558 & 0.474 & 0.425 & 0.400\\
    & \bm{$\uparrow$} \textbf{35.6\%} & \bm{$\uparrow$} \textbf{37.2\%} & \bm{$\uparrow$} \textbf{34.8\%} & \bm{$\uparrow$} \textbf{35.2\%} & \bm{$\uparrow$} \textbf{17.2\%} & \bm{$\uparrow$} \textbf{24.7\%} & \bm{$\uparrow$} \textbf{27.1\%} & \bm{$\uparrow$} \textbf{29.1\%} & \bm{$\uparrow$} \textbf{20.0\%} & \bm{$\uparrow$} \textbf{13.1\%} & \bm{$\uparrow$} \textbf{10.0\%} & \bm{$\uparrow$} \textbf{12.5\%} & \bm{$\uparrow$} \textbf{6.9\%} & \bm{$\uparrow$} \textbf{10.0\%} & \bm{$\uparrow$} \textbf{12.4\%} & \bm{$\uparrow$} \textbf{18.7\%}\\
    \cmidrule(l){1-17}
    SCHOLAR & 0.513 & 0.441 & 0.411 & 0.384 & 0.278 & 0.197 & 0.158 & 0.136 & 0.587 & 0.213 & 0.134 & 0.099 & 0.548 & 0.351 & 0.231 & 0.142 \\ 
    + LLM-ITL & 0.597 & 0.510 & 0.463 & 0.413 & 0.306 & 0.208 & 0.168 & 0.141 & 0.628 & 0.293 & 0.222 & 0.184 & 0.591 & 0.389 & 0.296 & 0.183 \\
    & \bm{$\uparrow$} \textbf{16.4\%} & \bm{$\uparrow$} \textbf{15.6\%} & \bm{$\uparrow$} \textbf{12.7\%} & \bm{$\uparrow$} \textbf{7.6\%} & \bm{$\uparrow$} \textbf{10.1\%} & \bm{$\uparrow$} \textbf{5.6\%} & \bm{$\uparrow$} \textbf{6.3\%} & \bm{$\uparrow$} \textbf{3.7\%} & \bm{$\uparrow$} \textbf{7.0\%} & \bm{$\uparrow$} \textbf{37.6\%} & \bm{$\uparrow$} \textbf{65.7\%} & \bm{$\uparrow$} \textbf{85.9\%} & \bm{$\uparrow$} \textbf{7.8\%} & \bm{$\uparrow$} \textbf{10.8\%} & \bm{$\uparrow$} \textbf{28.1\%} & \bm{$\uparrow$} \textbf{28.9\%}\\
    \cmidrule(l){1-17}
    ETM  & 0.455 & 0.387 & 0.320 & 0.278 & 0.307 & 0.225 & 0.192 & 0.161 & 0.466 & 0.438 & 0.381 & 0.320 & 0.498 & 0.412 & 0.367 & 0.320 \\
    + LLM-ITL & 0.589 & 0.481 & 0.427 & 0.385 & 0.385 & 0.296 & 0.249 & 0.219 & 0.686 & 0.597 & 0.528 & 0.482 & 0.585 & 0.458 & 0.395 & 0.357 \\
    & \bm{$\uparrow$} \textbf{29.5\%} & \bm{$\uparrow$} \textbf{24.3\%} & \bm{$\uparrow$} \textbf{33.4\%} & \bm{$\uparrow$} \textbf{38.5\%} & \bm{$\uparrow$} \textbf{25.4\%} & \bm{$\uparrow$} \textbf{31.6\%} & \bm{$\uparrow$} \textbf{29.7\%} & \bm{$\uparrow$} \textbf{36.0\%} & \bm{$\uparrow$} \textbf{47.2\%} & \bm{$\uparrow$} \textbf{36.3\%} & \bm{$\uparrow$} \textbf{38.6\%} & \bm{$\uparrow$} \textbf{50.6\%} & \bm{$\uparrow$} \textbf{17.5\%} & \bm{$\uparrow$} \textbf{11.2\%} & \bm{$\uparrow$} \textbf{7.6\%} & \bm{$\uparrow$} \textbf{11.6\%}  \\
    \cmidrule(l){1-17}
    NSTM & 0.291 & 0.226 & 0.164 & 0.140 & 0.191 & 0.132 & 0.099 & 0.091 & 0.557 & 0.488 & 0.421 & 0.383 & 0.489 & 0.443 & 0.408 & 0.404 \\
    + LLM-ITL & 0.367 & 0.309 & 0.295 & 0.266 & 0.245 & 0.171 & 0.161 & 0.119 & 0.607 & 0.515 & 0.470 & 0.416 & 0.569 & 0.495 & 0.490 & 0.421 \\
    & \bm{$\uparrow$} \textbf{26.1\%} & \bm{$\uparrow$} \textbf{36.7\%} & \bm{$\uparrow$} \textbf{79.9\%} & \bm{$\uparrow$} \textbf{90.0\%} & \bm{$\uparrow$} \textbf{28.3\%} & \bm{$\uparrow$} \textbf{29.5\%} & \bm{$\uparrow$} \textbf{62.6\%} & \bm{$\uparrow$} \textbf{30.8\%} & \bm{$\uparrow$} \textbf{9.0\%} & \bm{$\uparrow$} \textbf{5.5\%} & \bm{$\uparrow$} \textbf{11.6\%} & \bm{$\uparrow$} \textbf{8.6\%} & \bm{$\uparrow$} \textbf{16.4\%} & \bm{$\uparrow$} \textbf{11.7\%} & \bm{$\uparrow$} \textbf{20.1\%} & \bm{$\uparrow$} \textbf{4.2\%} \\
    \cmidrule(l){1-17}
    CLNTM  & 0.490 & 0.445 & 0.406 & 0.371 & 0.253 & 0.162 & 0.123 & 0.111 & 0.425 & 0.188 & 0.115 & 0.088 & 0.496 & 0.222 & 0.154 & 0.125 \\
    + LLM-ITL & 0.584 & 0.518 & 0.457 & 0.417 & 0.312 & 0.213 & 0.163 & 0.142 & 0.515 & 0.273 & 0.193 & 0.168 & 0.537 & 0.262 & 0.188 & 0.180 \\
    & \bm{$\uparrow$} \textbf{19.2\%} & \bm{$\uparrow$} \textbf{16.4\%} & \bm{$\uparrow$} \textbf{12.6\%} & \bm{$\uparrow$} \textbf{12.4\%} & \bm{$\uparrow$} \textbf{23.3\%} & \bm{$\uparrow$} \textbf{31.5\%} & \bm{$\uparrow$} \textbf{32.5\%} & \bm{$\uparrow$} \textbf{27.9\%} & \bm{$\uparrow$} \textbf{21.2\%} & \bm{$\uparrow$} \textbf{45.2\%} & \bm{$\uparrow$} \textbf{67.8\%} & \bm{$\uparrow$} \textbf{90.9\%} & \bm{$\uparrow$} \textbf{8.3\%} & \bm{$\uparrow$} \textbf{18.0\%} & \bm{$\uparrow$} \textbf{22.1\%} & \bm{$\uparrow$} \textbf{44.0\%}  \\
    \cmidrule(l){1-17}
    WeTe  & 0.375 & 0.216 & 0.193 & 0.154 & 0.435 & 0.326 & 0.249 & 0.203 & 0.519 & 0.418 & 0.326 & 0.294 & 0.538 & 0.514 & 0.399 & 0.384 \\
    + LLM-ITL & 0.474 & 0.359 & 0.310 & 0.249 & 0.504 & 0.365 & 0.295 & 0.255 & 0.583 & 0.530 & 0.487 & 0.412 & 0.587 & 0.571 & 0.508 & 0.512 \\
    &  \bm{$\uparrow$} \textbf{26.4\%} & \bm{$\uparrow$} \textbf{66.2\%} & \bm{$\uparrow$} \textbf{60.6\%} & \bm{$\uparrow$} \textbf{61.7\%} & \bm{$\uparrow$} \textbf{15.9\%} & \bm{$\uparrow$} \textbf{12.0\%} & \bm{$\uparrow$} \textbf{18.5\%} & \bm{$\uparrow$} \textbf{25.6\%} & \bm{$\uparrow$} \textbf{12.3\%} & \bm{$\uparrow$} \textbf{26.8\%} & \bm{$\uparrow$} \textbf{49.4\%} & \bm{$\uparrow$} \textbf{40.1\%} & \bm{$\uparrow$} \textbf{9.1\%} & \bm{$\uparrow$} \textbf{11.1\%} & \bm{$\uparrow$} \textbf{27.3\%} & \bm{$\uparrow$} \textbf{33.3\%} \\
    \cmidrule(l){1-17}
    ECRTM  &  0.343 & 0.319 & 0.312 & 0.302 & 0.273 & 0.299 & 0.278 & 0.281 & 0.561 & 0.486 & 0.424 & 0.401 & 0.438 & 0.408 & 0.381 & 0.352 \\
    + LLM-ITL & 0.501 & 0.412 & 0.363 & 0.328 & 0.336 & 0.349 & 0.336 & 0.337 & 0.633 & 0.565 & 0.521 & 0.486 & 0.505 & 0.464 & 0.430 & 0.396 \\
    &  \bm{$\uparrow$} \textbf{46.1\%} & \bm{$\uparrow$} \textbf{29.2\%} & \bm{$\uparrow$} \textbf{16.3\%} & \bm{$\uparrow$} \textbf{8.6\%} & \bm{$\uparrow$} \textbf{23.1\%} & \bm{$\uparrow$} \textbf{16.7\%} & \bm{$\uparrow$} \textbf{20.9\%} & \bm{$\uparrow$} \textbf{19.9\%} & \bm{$\uparrow$} \textbf{12.8\%} & \bm{$\uparrow$} \textbf{16.3\%} & \bm{$\uparrow$} \textbf{22.9\%} & \bm{$\uparrow$} \textbf{21.2\%} & \bm{$\uparrow$} \textbf{15.3\%} & \bm{$\uparrow$} \textbf{13.7\%} & \bm{$\uparrow$} \textbf{12.9\%} & \bm{$\uparrow$} \textbf{12.5\%} \\
    \bottomrule
    \end{tabular}
    }
    \caption{Topic quality ($\text{TQ}:=C_V\times \text{TD}$) at different $K$. Std. values have been omitted to save space in this table.} \label{tab:tq_k}
\end{table*}

\newpage

\begin{table*}[!t]
    \centering
    \resizebox{\textwidth}{!}{
    \begin{tabular}{lcccc|cccc|cccc|cccc}
    \toprule
    \multirow{2}{*}{\textbf{Model}} & \multicolumn{4}{c}{\textbf{20News}} & \multicolumn{4}{c}{\textbf{R8}} & \multicolumn{4}{c}{\textbf{DBpedia}} & \multicolumn{4}{c}{\textbf{AGNews}}\\
    \cmidrule(l){2-5} \cmidrule(l){6-9} \cmidrule(l){10-13} \cmidrule(l){14-17} 
    & \textbf{25} & \textbf{50} & \textbf{75} & \textbf{100} & \textbf{25} & \textbf{50} & \textbf{75} & \textbf{100} &  \textbf{25} & \textbf{50} & \textbf{75} & \textbf{100} & \textbf{25} & \textbf{50} & \textbf{75} & \textbf{100}\\
    \midrule
    LDA  & 0.457 & 0.521 & 0.527 & 0.531 & 0.907 & 0.920 & 0.927 & 0.926 & 0.813 & 0.833 & 0.837 & 0.839 & 0.818 & 0.821 & 0.822 & 0.822\\
    BERTopic  & 0.313 & 0.371 & 0.404 & 0.425 & 0.882 & 0.875 & 0.892 & 0.897 & 0.748 & 0.817 & 0.810 & 0.795 & 0.648 & 0.688 & 0.716 & 0.738\\
    TopicGPT  & 0.336 & 0.336 & 0.336 & 0.336 & 0.577 & 0.577 & 0.577 & 0.577 & 0.718 & 0.718 & 0.718 & 0.718 & 0.819 & 0.819 & 0.819 & 0.819\\
    \cmidrule(l){1-17}
    NVDM  & 0.166 & 0.165 & 0.166 & 0.170 & 0.610 & 0.600 & 0.596 & 0.603 & 0.210 & 0.193 & 0.199 & 0.204 & 0.432 & 0.402 & 0.389 & 0.389 \\
    + LLM-ITL & 0.162 & 0.163 & 0.166 & 0.172 & 0.610 & 0.603 & 0.598 & 0.602 & 0.206 & 0.195 & 0.197 & 0.200 & 0.428 & 0.401 & 0.388 & 0.389\\
    & $\downarrow$ 2.4\% & $\downarrow$ 1.2\% & \bm{$\uparrow$} \textbf{0.0\%} & \bm{$\uparrow$} \textbf{1.2\%} & \bm{$\uparrow$} \textbf{0.0\%} & \bm{$\uparrow$} \textbf{0.5\%} & \bm{$\uparrow$} \textbf{0.3\%} & $\downarrow$ 0.2\% & $\downarrow$ 1.9\% & \bm{$\uparrow$} \textbf{1.0\%} & $\downarrow$ 1.0\% & $\downarrow$ 2.0\% & $\downarrow$ 0.9\% & $\downarrow$ 0.2\% & $\downarrow$ 0.3\% & \bm{$\uparrow$} \textbf{0.0\%} \\
     \cmidrule(l){1-17}
    PLDA  & 0.272 & 0.317 & 0.337 & 0.346 & 0.756 & 0.770 & 0.775 & 0.777 & 0.746 & 0.749 & 0.733 & 0.730 & 0.716 & 0.727 & 0.731 & 0.730 \\
    + LLM-ITL & 0.284 & 0.318 & 0.355 & 0.361 & 0.763 & 0.767 & 0.771 & 0.771 & 0.742 & 0.746 & 0.739 & 0.732 & 0.722 & 0.725 & 0.727 & 0.726 \\
    & \bm{$\uparrow$} \textbf{4.4\%} & \bm{$\uparrow$} \textbf{0.3\%} & \bm{$\uparrow$} \textbf{5.3\%} & \bm{$\uparrow$} \textbf{4.3\%} & \bm{$\uparrow$} \textbf{0.9\%} & $\downarrow$ 0.4\% & $\downarrow$ 0.5\% & $\downarrow$ 0.8\% & $\downarrow$ 0.5\% & $\downarrow$ 0.4\% & \bm{$\uparrow$} \textbf{0.8\%} & \bm{$\uparrow$} \textbf{0.3\%} & \bm{$\uparrow$} \textbf{0.8\%} & $\downarrow$ 0.3\% & $\downarrow$ 0.5\% & $\downarrow$ 0.5\%\\
    \cmidrule(l){1-17}
    SCHOLAR & 0.597 & 0.627 & 0.634 & 0.650 & 0.921 & 0.911 & 0.930 & 0.928 & 0.872 & 0.864 & 0.822 & 0.690 & 0.855 & 0.790 & 0.570 & 0.533 \\ 
    + LLM-ITL & 0.580 & 0.607 & 0.631 & 0.649 & 0.920 & 0.911 & 0.927 & 0.923 & 0.875 & 0.865 & 0.798 & 0.709 & 0.855 & 0.765 & 0.546 & 0.523\\
    & $\downarrow$ 2.8\% & $\downarrow$ 3.2\% & $\downarrow$ 0.5\% & $\downarrow$ 0.2\% & $\downarrow$ 0.1\% & \bm{$\uparrow$} \textbf{0.0\%} & $\downarrow$ 0.3\% & $\downarrow$ 0.5\% & \bm{$\uparrow$} \textbf{0.3\%} & \bm{$\uparrow$} \textbf{0.1\%} & $\downarrow$ 2.9\% & \bm{$\uparrow$} \textbf{2.8\%} & \bm{$\uparrow$} \textbf{0.0\%} & $\downarrow$ 3.2\% & $\downarrow$ 4.2\% & $\downarrow$ 1.9\% \\
    \cmidrule(l){1-17}
    ETM  & 0.355 & 0.410 & 0.444 & 0.471 & 0.872 & 0.875 & 0.887 & 0.877 & 0.794 & 0.813 & 0.817 & 0.812 & 0.783 & 0.781 & 0.789 & 0.787 \\
    + LLM-ITL & 0.348 & 0.403 & 0.436 & 0.457 & 0.860 & 0.875 & 0.870 & 0.861 & 0.760 & 0.781 & 0.798 & 0.811 & 0.784 & 0.784 & 0.794 & 0.790 \\
    & $\downarrow$ 2.0\% & $\downarrow$ 1.7\% & $\downarrow$ 1.8\% & $\downarrow$ 3.0\% & $\downarrow$ 1.4\% & \bm{$\uparrow$} \textbf{0.0\%} & $\downarrow$ 1.9\% & $\downarrow$ 1.8\% & $\downarrow$ 4.3\% & $\downarrow$ 3.9\% & $\downarrow$ 2.3\% & $\downarrow$ 0.1\% & \bm{$\uparrow$} \textbf{0.1\%} & \bm{$\uparrow$} \textbf{0.4\%} & \bm{$\uparrow$} \textbf{0.6\%} & \bm{$\uparrow$} \textbf{0.4\%} \\
    \cmidrule(l){1-17}
    NSTM & 0.356 & 0.375 & 0.390 & 0.423 & 0.883 & 0.895 & 0.903 & 0.908 & 0.818 & 0.825 & 0.840 & 0.848 & 0.818 & 0.819 & 0.821 & 0.825 \\
    + LLM-ITL & 0.341 & 0.361 & 0.371 & 0.395 & 0.884 & 0.896 & 0.903 & 0.905 & 0.797 & 0.825 & 0.836 & 0.843 & 0.807 & 0.819 & 0.817 & 0.822\\
    & $\downarrow$ 4.2\% & $\downarrow$ 3.7\% & $\downarrow$ 4.9\% & $\downarrow$ 6.6\% & \bm{$\uparrow$} \textbf{0.1\%} & \bm{$\uparrow$} \textbf{0.1\%} & \bm{$\uparrow$} \textbf{0.0\%} & $\downarrow$ 0.3\% & $\downarrow$ 2.6\% & \bm{$\uparrow$} \textbf{0.0\%} & $\downarrow$ 0.5\% & $\downarrow$ 0.6\% & $\downarrow$ 1.3\% & \bm{$\uparrow$} \textbf{0.0\%} & $\downarrow$ 0.5\% & $\downarrow$ 0.4\%\\
    \cmidrule(l){1-17}
    CLNTM  & 0.551 & 0.623 & 0.637 & 0.650 & 0.894 & 0.923 & 0.926 & 0.929 & 0.725 & 0.574 & 0.462 & 0.454 & 0.821 & 0.766 & 0.641 & 0.564 \\
    + LLM-ITL & 0.543 & 0.623 & 0.645 & 0.649 & 0.893 & 0.923 & 0.925 & 0.928 & 0.727 & 0.565 & 0.461 & 0.452 & 0.806 & 0.765 & 0.616 & 0.591 \\
    & $\downarrow$ 1.5\% & \bm{$\uparrow$} \textbf{0.0\%} & \bm{$\uparrow$} \textbf{1.3\%} & $\downarrow$ 0.2\% & $\downarrow$ 0.1\% & \bm{$\uparrow$} \textbf{0.0\%} & $\downarrow$ 0.1\% & $\downarrow$ 0.1\% & \bm{$\uparrow$} \textbf{0.3\%} & $\downarrow$ 1.6\% & $\downarrow$ 0.2\% & $\downarrow$ 0.4\% & $\downarrow$ 1.8\% & $\downarrow$ 0.1\% & $\downarrow$ 3.9\% & \bm{$\uparrow$} \textbf{4.8\%} \\
    \cmidrule(l){1-17}
    WeTe  & 0.284 & 0.327 & 0.347 & 0.360 & 0.868 & 0.860 & 0.870 & 0.867 & 0.739 & 0.765 & 0.784 & 0.796 & 0.722 & 0.729 & 0.735 & 0.724 \\
    + LLM-ITL & 0.281 & 0.327 & 0.352 & 0.356 & 0.872 & 0.862 & 0.878 & 0.872 & 0.718 & 0.761 & 0.790 & 0.804 & 0.725 & 0.732 & 0.718 & 0.741\\
    & $\downarrow$ 1.1\% & \bm{$\uparrow$} \textbf{0.0\%} & \bm{$\uparrow$} \textbf{1.4\%} & $\downarrow$ 1.1\% & \bm{$\uparrow$} \textbf{0.5\%} & \bm{$\uparrow$} \textbf{0.2\%} & \bm{$\uparrow$} \textbf{0.9\%} & \bm{$\uparrow$} \textbf{0.6\%} & $\downarrow$ 2.8\% & $\downarrow$ 0.5\% & \bm{$\uparrow$} \textbf{0.8\%} & \bm{$\uparrow$} \textbf{1.0\%} & \bm{$\uparrow$} \textbf{0.4\%} & \bm{$\uparrow$} \textbf{0.4\%} & $\downarrow$ 2.3\% & \bm{$\uparrow$} \textbf{2.3\%} \\
    \cmidrule(l){1-17}
    ECRTM  & 0.549 & 0.552 & 0.574 & 0.579 & 0.837 & 0.886 & 0.893 & 0.891 & 0.715 & 0.824 & 0.808 & 0.833 & 0.765 & 0.771 & 0.763 & 0.764 \\
    + LLM-ITL & 0.549 & 0.555 & 0.565 & 0.574 & 0.834 & 0.881 & 0.890 & 0.894 & 0.747 & 0.827 & 0.840 & 0.830 & 0.765 & 0.769 & 0.761 & 0.725 \\
    & \bm{$\uparrow$} \textbf{0.0\%} & \bm{$\uparrow$} \textbf{0.5\%} & $\downarrow$ 1.6\% & $\downarrow$ 0.9\% & $\downarrow$ 0.4\% & $\downarrow$ 0.6\% & $\downarrow$ 0.3\% & \bm{$\uparrow$} \textbf{0.3\%} & \bm{$\uparrow$} \textbf{4.5\%} & \bm{$\uparrow$} \textbf{0.4\%} & \bm{$\uparrow$} \textbf{4.0\%} & $\downarrow$ 0.4\% & \bm{$\uparrow$} \textbf{0.0\%} & $\downarrow$ 0.3\% & $\downarrow$ 0.3\% & $\downarrow$ 5.1\% \\
    \bottomrule
    \end{tabular}
    }
    \caption{Purity at different $K$. Std. values have been omitted to save space in this table.} \label{tab:tp_k}
\end{table*}

\hfill

\begin{table*}[!h]
    \centering
    \resizebox{\textwidth}{!}{
    \begin{tabular}{lcccc|cccc|cccc|cccc}
    \toprule
    \multirow{2}{*}{\textbf{Model}} & \multicolumn{4}{c}{\textbf{20News}} & \multicolumn{4}{c}{\textbf{R8}} & \multicolumn{4}{c}{\textbf{DBpedia}} & \multicolumn{4}{c}{\textbf{AGNews}}\\
    \cmidrule(l){2-5} \cmidrule(l){6-9} \cmidrule(l){10-13} \cmidrule(l){14-17} 
    & \textbf{25} & \textbf{50} & \textbf{75} & \textbf{100} & \textbf{25} & \textbf{50} & \textbf{75} & \textbf{100} &  \textbf{25} & \textbf{50} & \textbf{75} & \textbf{100} & \textbf{25} & \textbf{50} & \textbf{75} & \textbf{100}\\
    \midrule
    LDA  & 0.463 & 0.456 & 0.449 & 0.447 & 0.521 & 0.480 & 0.472 & 0.452 & 0.712 & 0.662 & 0.639 & 0.620 & 0.373 & 0.332 & 0.318 & 0.303 \\
    BERTopic  & 0.266 & 0.312 & 0.341 & 0.355 & 0.582 & 0.519 & 0.486 & 0.474 & 0.691 & 0.662 & 0.622 & 0.595 & 0.252 & 0.255 & 0.252 & 0.251 \\
    TopicGPT  & 0.390 & 0.390 & 0.390 & 0.390 & 0.244 & 0.244 & 0.244 & 0.244 & 0.694 & 0.694 & 0.694 & 0.694 & 0.449 & 0.449 & 0.449 & 0.449\\
    \cmidrule(l){1-17}
    NVDM  & 0.126 & 0.117 & 0.124 & 0.136 & 0.149 & 0.118 & 0.111 & 0.115 & 0.140 & 0.107 & 0.104 & 0.107 & 0.071 & 0.045 & 0.035 & 0.033 \\
    + LLM-ITL & 0.123 & 0.113 & 0.119 & 0.135 & 0.145 & 0.117 & 0.111 & 0.114 & 0.134 & 0.104 & 0.102 & 0.104 & 0.069 & 0.044 & 0.035 & 0.034\\
    & $\downarrow$ 2.4\% & $\downarrow$ 3.4\% & $\downarrow$ 4.0\% & $\downarrow$ 0.7\% & $\downarrow$ 2.7\% & $\downarrow$ 0.8\% & \bm{$\uparrow$} \textbf{0.0\%} & $\downarrow$ 0.9\% & $\downarrow$ 4.3\% & $\downarrow$ 2.8\% & $\downarrow$ 1.9\% & $\downarrow$ 2.8\% & $\downarrow$ 2.8\% & $\downarrow$ 2.2\% & \bm{$\uparrow$} \textbf{0.0\%} & \bm{$\uparrow$} \textbf{3.0\%} \\
     \cmidrule(l){1-17}
    PLDA  & 0.191 & 0.228 & 0.243 & 0.260 & 0.291 & 0.279 & 0.265 & 0.262 & 0.582 & 0.537 & 0.506 & 0.491 & 0.246 & 0.222 & 0.219 & 0.212 \\
    + LLM-ITL & 0.199 & 0.225 & 0.255 & 0.266 & 0.303 & 0.277 & 0.264 & 0.258 & 0.578 & 0.536 & 0.509 & 0.489 & 0.249 & 0.220 & 0.214 & 0.207\\
    & \bm{$\uparrow$} \textbf{4.2\%} & $\downarrow$ 1.3\% & \bm{$\uparrow$} \textbf{4.9\%} & \bm{$\uparrow$} \textbf{2.3\%} & \bm{$\uparrow$} \textbf{4.1\%} & $\downarrow$ 0.7\% & $\downarrow$ 0.4\% & $\downarrow$ 1.5\% & $\downarrow$ 0.7\% & $\downarrow$ 0.2\% & \bm{$\uparrow$} \textbf{0.6\%} & $\downarrow$ 0.4\% & \bm{$\uparrow$} \textbf{1.2\%} & $\downarrow$ 0.9\% & $\downarrow$ 2.3\% & $\downarrow$ 2.4\% \\
    \cmidrule(l){1-17}
    SCHOLAR & 0.547 & 0.538 & 0.525 & 0.525 & 0.512 & 0.449 & 0.443 & 0.433 & 0.778 & 0.790 & 0.766 & 0.671 & 0.420 & 0.383 & 0.282 & 0.244 \\ 
    + LLM-ITL & 0.545 & 0.529 & 0.525 & 0.525 & 0.511 & 0.449 & 0.438 & 0.426 & 0.781 & 0.791 & 0.746 & 0.692 & 0.423 & 0.368 & 0.274 & 0.241 \\
    & $\downarrow$ 0.4\% & $\downarrow$ 1.7\% & \bm{$\uparrow$} \textbf{0.0\%} & \bm{$\uparrow$} \textbf{0.0\%} & $\downarrow$ 0.2\% & \bm{$\uparrow$} \textbf{0.0\%} & $\downarrow$ 1.1\% & $\downarrow$ 1.6\% & \bm{$\uparrow$} \textbf{0.4\%} & \bm{$\uparrow$} \textbf{0.1\%} & $\downarrow$ 2.6\% & \bm{$\uparrow$} \textbf{3.1\%} & \bm{$\uparrow$} \textbf{0.7\%} & $\downarrow$ 3.9\% & $\downarrow$ 2.8\% & $\downarrow$ 1.2\%  \\
    \cmidrule(l){1-17}
    ETM  & 0.371 & 0.405 & 0.424 & 0.442 & 0.485 & 0.463 & 0.460 & 0.438 & 0.728 & 0.690 & 0.678 & 0.667 & 0.352 & 0.325 & 0.321 & 0.311 \\
    + LLM-ITL & 0.357 & 0.394 & 0.416 & 0.425 & 0.519 & 0.498 & 0.476 & 0.471 & 0.723 & 0.698 & 0.678 & 0.676 & 0.354 & 0.332 & 0.327 & 0.315 \\
    & $\downarrow$ 3.8\% & $\downarrow$ 2.7\% & $\downarrow$ 1.9\% & $\downarrow$ 3.8\% & \bm{$\uparrow$} \textbf{7.0\%} & \bm{$\uparrow$} \textbf{7.6\%} & \bm{$\uparrow$} \textbf{3.5\%} & \bm{$\uparrow$} \textbf{7.5\%} & $\downarrow$ 0.7\% & \bm{$\uparrow$} \textbf{1.2\%} & \bm{$\uparrow$} \textbf{0.0\%} & \bm{$\uparrow$} \textbf{1.3\%} & \bm{$\uparrow$} \textbf{0.6\%} & \bm{$\uparrow$} \textbf{2.2\%} & \bm{$\uparrow$} \textbf{1.9\%} & \bm{$\uparrow$} \textbf{1.3\%} \\
    \cmidrule(l){1-17}
    NSTM & 0.375 & 0.371 & 0.375 & 0.388 & 0.465 & 0.435 & 0.422 & 0.415 & 0.717 & 0.653 & 0.638 & 0.628 & 0.368 & 0.334 & 0.319 & 0.313 \\
    + LLM-ITL & 0.362 & 0.359 & 0.362 & 0.373 & 0.477 & 0.450 & 0.435 & 0.433 & 0.706 & 0.656 & 0.639 & 0.630 & 0.362 & 0.333 & 0.316 & 0.311\\
    & $\downarrow$ 3.5\% & $\downarrow$ 3.2\% & $\downarrow$ 3.5\% & $\downarrow$ 3.9\% & \bm{$\uparrow$} \textbf{2.6\%} & \bm{$\uparrow$} \textbf{3.4\%} & \bm{$\uparrow$} \textbf{3.1\%} & \bm{$\uparrow$} \textbf{4.3\%} & $\downarrow$ 1.5\% & \bm{$\uparrow$} \textbf{0.5\%} & \bm{$\uparrow$} \textbf{0.2\%} & \bm{$\uparrow$} \textbf{0.3\%} & $\downarrow$ 1.6\% & $\downarrow$ 0.3\% & $\downarrow$ 0.9\% & $\downarrow$ 0.6\%\\
    \cmidrule(l){1-17}
    CLNTM  & 0.522 & 0.526 & 0.524 & 0.521 & 0.489 & 0.459 & 0.434 & 0.427 & 0.641 & 0.510 & 0.414 & 0.400 & 0.392 & 0.390 & 0.333 & 0.311 \\
    + LLM-ITL & 0.514 & 0.529 & 0.525 & 0.521 & 0.486 & 0.459 & 0.434 & 0.425 & 0.640 & 0.500 & 0.415 & 0.403 & 0.382 & 0.394 & 0.321 & 0.315 \\
    & $\downarrow$ 1.5\% & \bm{$\uparrow$} \textbf{0.6\%} & \bm{$\uparrow$} \textbf{0.2\%} & \bm{$\uparrow$} \textbf{0.0\%} & $\downarrow$ 0.6\% & \bm{$\uparrow$} \textbf{0.0\%} & \bm{$\uparrow$} \textbf{0.0\%} & $\downarrow$ 0.5\% & $\downarrow$ 0.2\% & $\downarrow$ 2.0\% & \bm{$\uparrow$} \textbf{0.2\%} & \bm{$\uparrow$} \textbf{0.8\%} & $\downarrow$ 2.6\% & \bm{$\uparrow$} \textbf{1.0\%} & $\downarrow$ 3.6\% & \bm{$\uparrow$} \textbf{1.3\%} \\
    \cmidrule(l){1-17}
    WeTe  & 0.285 & 0.309 & 0.311 & 0.319 & 0.488 & 0.440 & 0.471 & 0.431 & 0.702 & 0.683 & 0.667 & 0.653 & 0.310 & 0.275 & 0.270 & 0.250 \\
    + LLM-ITL & 0.276 & 0.305 & 0.318 & 0.315 & 0.498 & 0.449 & 0.479 & 0.434 & 0.688 & 0.683 & 0.663 & 0.650 & 0.310 & 0.271 & 0.260 & 0.259\\
    & $\downarrow$ 3.2\% & $\downarrow$ 1.3\% & \bm{$\uparrow$} \textbf{2.3\%} & $\downarrow$ 1.3\% & \bm{$\uparrow$} \textbf{2.0\%} & \bm{$\uparrow$} \textbf{2.0\%} & \bm{$\uparrow$} \textbf{1.7\%} & \bm{$\uparrow$} \textbf{0.7\%} & $\downarrow$ 2.0\% & \bm{$\uparrow$} \textbf{0.0\%} & $\downarrow$ 0.6\% & $\downarrow$ 0.5\% & \bm{$\uparrow$} \textbf{0.0\%} & $\downarrow$ 1.5\% & $\downarrow$ 3.7\% & \bm{$\uparrow$} \textbf{3.6\%} \\
    \cmidrule(l){1-17}
    ECRTM  & 0.492 & 0.480 & 0.492 & 0.492 & 0.467 & 0.459 & 0.448 & 0.439 & 0.621 & 0.677 & 0.682 & 0.712 & 0.342 & 0.316 & 0.309 & 0.314 \\
    + LLM-ITL & 0.490 & 0.482 & 0.489 & 0.491 & 0.455 & 0.452 & 0.442 & 0.439 & 0.645 & 0.689 & 0.712 & 0.721 & 0.341 & 0.313 & 0.317 & 0.322\\
    & $\downarrow$ 0.4\% & \bm{$\uparrow$} \textbf{0.4\%} & $\downarrow$ 0.6\% & $\downarrow$ 0.2\% & $\downarrow$ 2.6\% & $\downarrow$ 1.5\% & $\downarrow$ 1.3\% & \bm{$\uparrow$} \textbf{0.0\%} & \bm{$\uparrow$} \textbf{3.9\%} & \bm{$\uparrow$} \textbf{1.8\%} & \bm{$\uparrow$} \textbf{4.4\%} & \bm{$\uparrow$} \textbf{1.3\%} & $\downarrow$ 0.3\% & $\downarrow$ 0.9\% & \bm{$\uparrow$} \textbf{2.6\%} & \bm{$\uparrow$} \textbf{2.5\%} \\
    \bottomrule
    \end{tabular}
    }
    \caption{NMI at different $K$. Std. values have been omitted to save space in this table.} \label{tab:tn_k}
\end{table*}

\newpage

\twocolumn

\section{Study of Confidence Alternatives}\label{appendix:conf}
\vspace{-5mm}
\setcounter{figure}{0}
\renewcommand{\thefigure}{F\arabic{figure}}
\begin{figure}[H]
    \centering
    \includegraphics[width=0.5\textwidth]{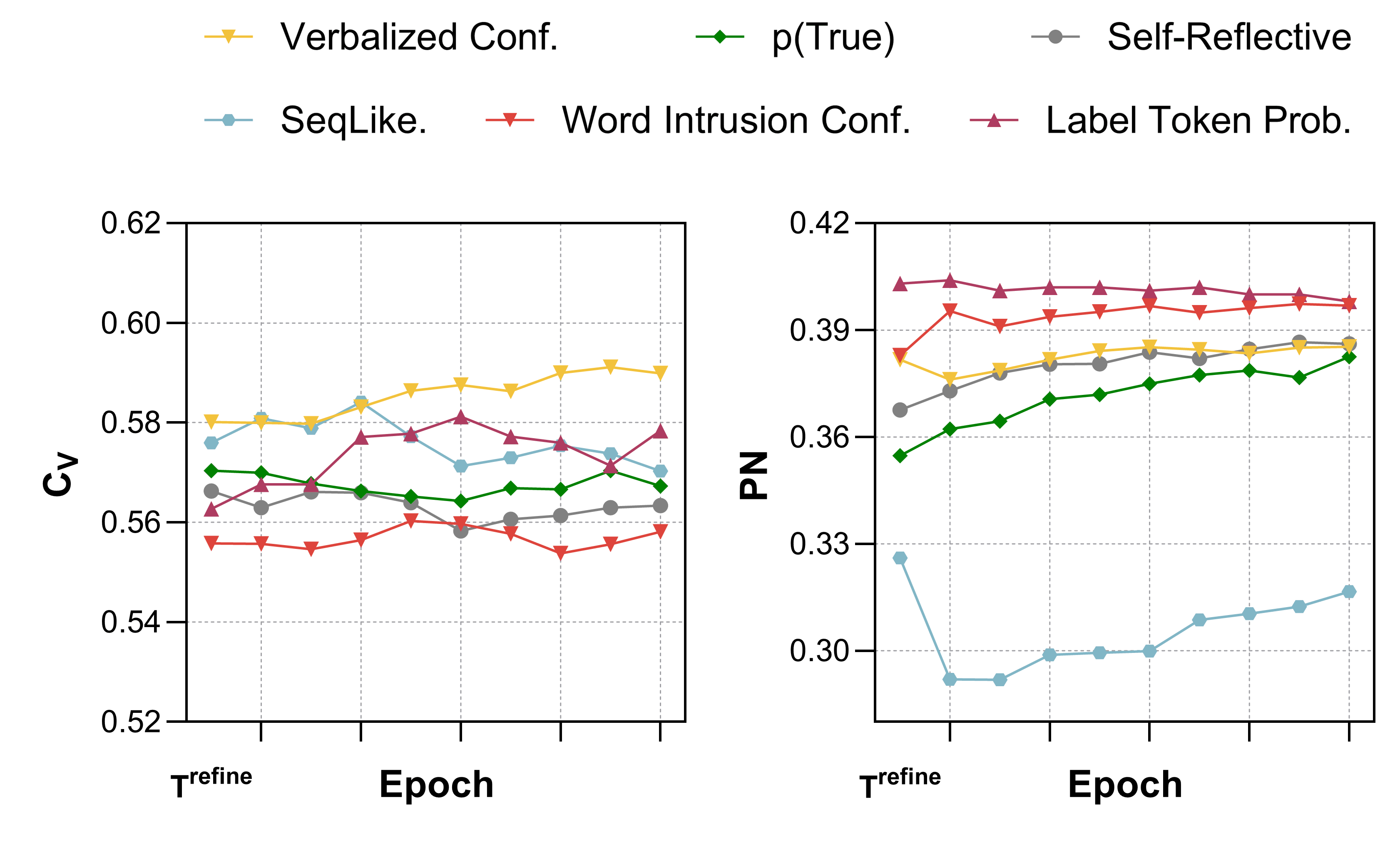}
    \vspace{-5mm}
  \caption{Learning curves of LLM-ITL (ETM as the base model) with different LLM confidences in terms of topic coherence ($C_V$) and topic alignment (PN) on 20News. Error bars are omitted for clarity in the figure.}
  \label{fig:all_conf}
  \vspace{-3mm}
\end{figure}
Here, we compare our proposed topic labeling confidence with other LLM confidence alternatives within the LLM-ITL framework. We limit our focus to single-sample approaches, where a single round of LLM inference is performed for a given topic. This ensures they can be efficiently applied within the training of LLM-ITL. We consider the following confidence alternatives to our topic labeling confidence in this study: (1) \textbf{No Conf.}, where no LLM confidence estimation is included, and $\text{Conf}(\bm{w}_{k}^l)=1$ in Eq. \ref{reg_loss} during the training. (2) \textbf{Verbalized Confidence} \citep{xiong2024can}, which directly asks the LLM for its confidence in solving a problem. The prompt we used for eliciting verbalized confidence is shown in Figure \ref{fig:prompt_refine_conf}. (3) \textbf{Self-Reflective Confidence} \citep{chen-mueller-2024-quantifying}, which prompts the LLM to evaluate its own answer in a two-stage manner. The topic label is obtained in the first stage, and the LLM evaluates this answer in a follow-up question (Figure \ref{fig:prompt_self_reflective}). (4) \textbf{p(True)} \citep{kadavath2022language}, which is similar to self-reflective confidence, but asks a true/false question instead. It takes the token probability of the response as the confidence measure (Figure \ref{fig:prompt_self_reflective}). (5) \textbf{SeqLike.} \citep{ren2023outofdistribution}, which computes the length-normalized sequence likelihood of the output from the LLM. 

From the results in Figure \ref{fig:all_conf}, we can observe that while verbalized confidence results in greater improvements in topic coherence, it can bias the topics toward the LLM’s knowledge rather than the input corpus, leading to reduced topic alignment. In contrast, label token probability and word intrusion confidence consistently achieve the best topic alignment performance, indicating a stronger relevance of the topics to the corpus.

\section{Study of Hyper-parameters}\label{appendix:hyper}
\vspace{-1mm}
\setcounter{figure}{0}
\renewcommand{\thefigure}{G\arabic{figure}}
\begin{figure}[H]
    \centering
    \vspace{-5mm}
\includegraphics[width=0.5\textwidth]{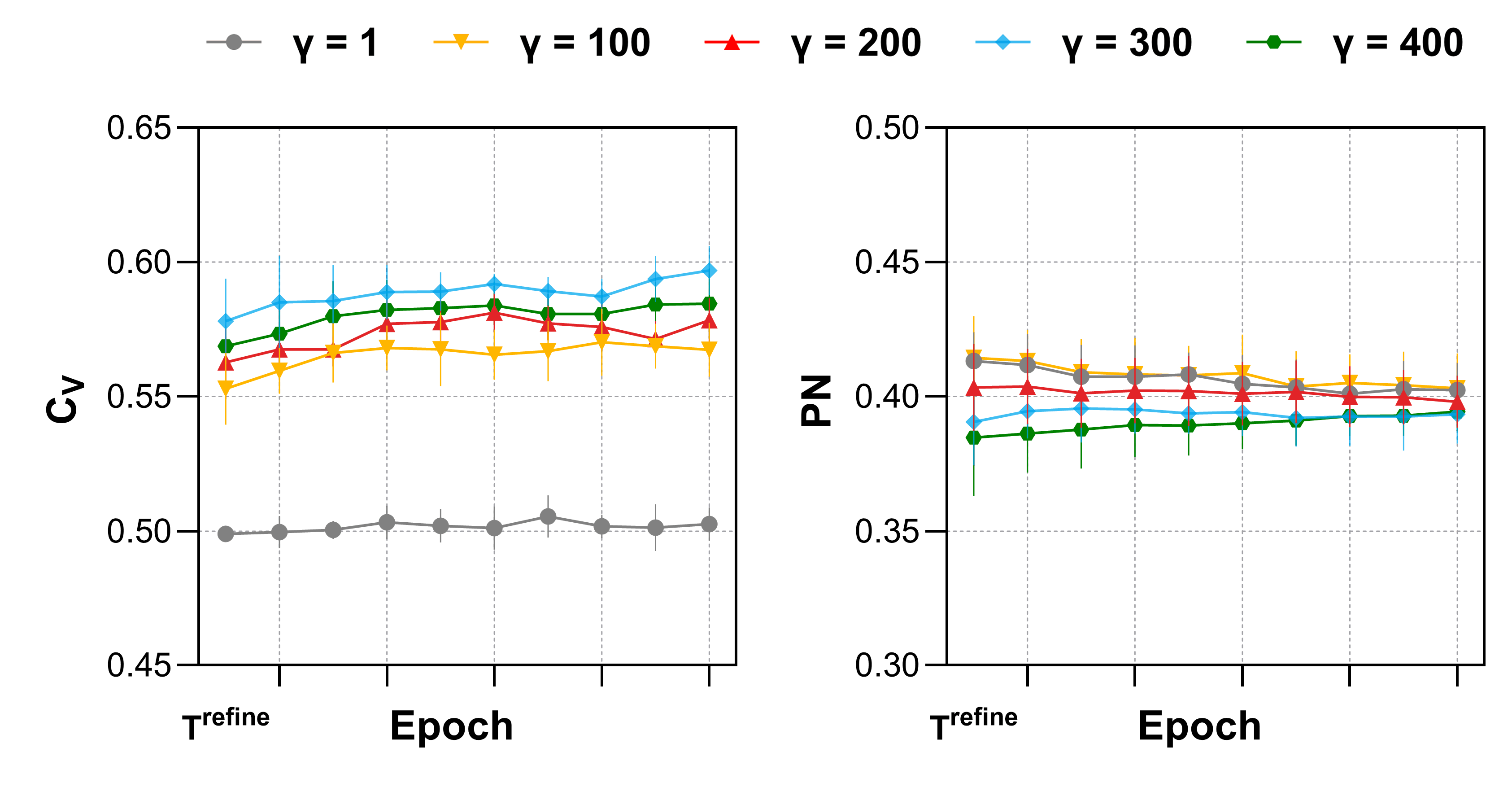}
  \caption{Learning curves of LLM-ITL (ETM as the base model) with different $\gamma$ in terms of topic coherence ($C_V$) and topic alignment (PN) on 20News.}
  \label{fig:hyper_weight}
\end{figure}
\begin{figure}[H]
    \centering
    \vspace{-5mm}
    \includegraphics[width=0.5\textwidth]{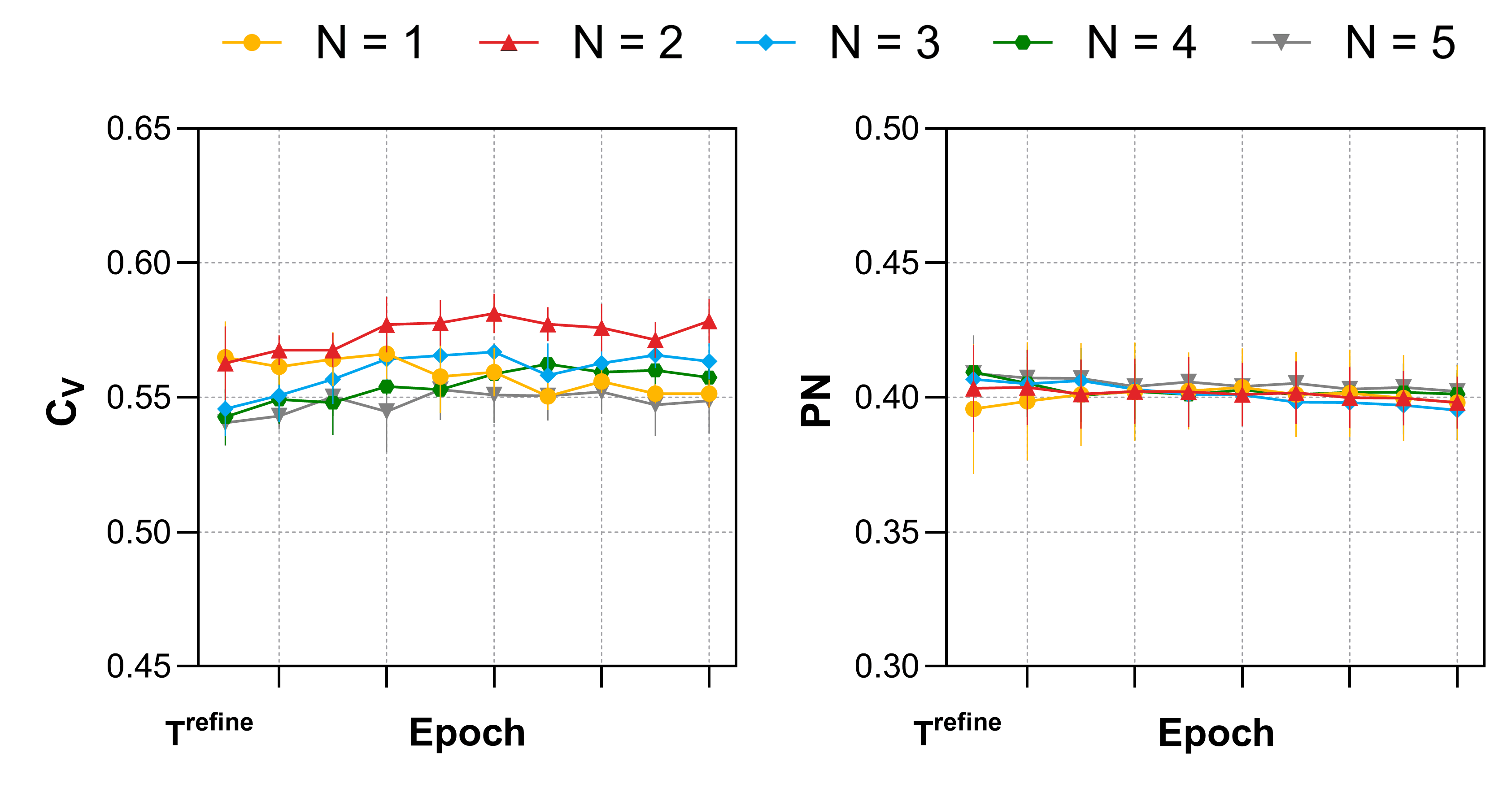}
  \caption{Learning curves of LLM-ITL (ETM as the base model) with different settings of $N$ (i.e., the number of words in the topic label) in terms of topic coherence ($C_V$) and topic alignment (PN) on 20News.}
  \label{fig:hyper_TG}
  \vspace{-3mm}
\end{figure}

Here, we study the hyper-parameter of LLM-ITL, focusing on topic refinement strength $\gamma$, and the number of words $N$ used for the topic label.

As for $\gamma$, we vary its value from 1 to 400 and plot the learning curves in terms of $C_V$ and PN, as shown in Figure \ref{fig:hyper_weight}. We can observe that: (1) In terms of topic coherence, $\gamma$ values between 100 and 300 yield  similar performance, suggesting low sensitivity to $\gamma$ within this range. (2) In terms of topic alignment, a higher $\gamma$ leads to slightly reduced performance in the initial phase. This occurs because relying heavily on topic refinement from the LLM causes the topics to bias towards the LLM's knowledge rather than the information from the input corpus. However, as training progresses, they converge to similar values. These observations suggest that LLM-ITL exhibits low sensitivity to $\gamma$ within a certain range and offers flexibility in controlling the balance between learning from the corpus and the LLM.

\setcounter{table}{0} 
\renewcommand{\thetable}{H\arabic{table}}
\begin{table*}[t]
    \centering
    \resizebox{\textwidth}{!}{
    \begin{tabular}{lcccc}
    \toprule
    \textbf{Prompt} & \textbf{Success Rate} $(\uparrow)$ & \textbf{N\_Input} $(\downarrow)$ & \textbf{N\_Output} $(\downarrow)$ & \textbf{Refined TC} $(\uparrow)$\\
    \midrule
       Origin & 0.978 & 224.48 & 197.86 & \underline{0.554}\\
       Variant\_1 & 0.967 & 228.48 & \textbf{80.81} & 0.512\\
       Variant\_2 & 0.935 & 269.48 & 167.05 & 0.498\\
       Variant\_3 & \underline{0.980} & 223.48 & 172.51 & 0.537\\
       Variant\_4 & 0.972 & \textbf{189.48} & 191.04 & \textbf{0.562}\\
       Variant\_5 & 0.956 & \underline{208.48} & \underline{159.66} & 0.543\\
       Iterative Refinement \citep{chang2024enhanced} & \textbf{0.993} & 1872.13 & 936.70 & 0.478 \\
       \bottomrule
    \end{tabular}
    }
    \vspace{-2mm}
\caption{Study of prompt variants. The best and second-best performance of each column are highlighted in boldface and underlined, respectively.}
\vspace{-5mm}
\label{tab:prompt_performance}
\end{table*}

As for $N$, we vary the number of words for the topic label from 1 to 5 and plot the learning curves in terms of $C_V$ and PN. As illustrated in Figure \ref{fig:hyper_TG}, we can observe that: (1) Using more words as the topic label for LLM-ITL, such as a 5-word topic label (e.g., $N=5$), results in the least improvement in topic coherence. While using a 2-word topic label (e.g., $N=2$) achieves the best performance in terms of topic coherence. (2) For topic alignment, the number of words in the topic label demonstrates comparable performance, suggesting low sensitivity to $N$.

\vspace{-1mm}
\section{Study of Prompts}\label{appendix:prompt_variants}
\vspace{-2mm}
\setcounter{figure}{0}
\renewcommand{\thefigure}{H\arabic{figure}}

\paragraph{Prompt Variants} Here, we study the effectiveness of different prompts for topic suggestion. We obtain variants of topic suggestion prompts in Figure \ref{fig:prompt_refine} through modifying the technique used in PromptBreeder  \citep{10.5555/3692070.3692611}. To be specific, we first create a set of 100 ``mutation-prompts'' (e.g., ``Make a variant of the prompt") and 100 ``thinking-styles'' (e.g., ``Let’s think step by step"). We generate a set of 50 task prompts by concatenating a randomly drawn ``mutation-prompt'' and a randomly drawn ``thinking-style'' to the original prompt, and provide that to the Claude 3.5\footnote{https://www.anthropic.com/news/claude-3-family} to produce a continuation, resulting in a different task prompt. Secondly, we randomly select 50 topics over 4 experiment datasets. We run those topics through 50 generated task prompts and filter out the generated prompts that cannot give JSON format in the selected topics or generate above 300 tokens. We are left with 14 topics. We then leverage Claude 3.5 to judge the quality of generated topics and refined topic words. We rank 14 methods by overall topics and refined topic words to get 5 variants of prompts. In addition to the prompt variants generated by those steps, we also investigate the topic refinement prompt used in \cite{chang2024enhanced} (see Figure 2 of their paper). All the prompt variants for topic refinement in this study are illustrated in Table \ref{prompt_variants}. 

\begin{figure}[t]
    \centering
    \includegraphics[width=0.5\textwidth]{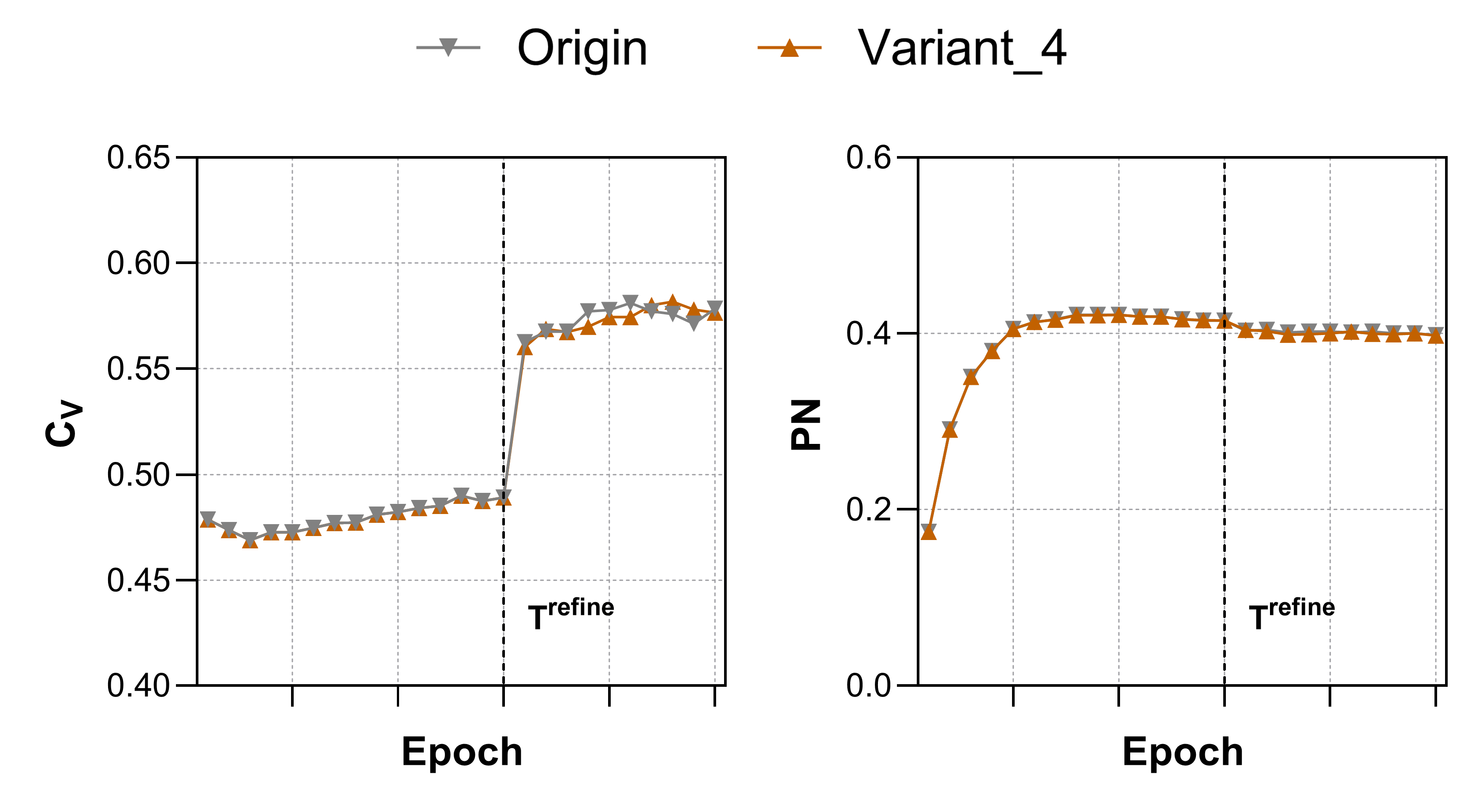}
    \vspace{-5mm}
  \caption{Learning curves of LLM-ITL (ETM as the base model) with different prompts in terms of $C_V$ and PN on 20News.}
  \label{fig:prompt_variants}
  \vspace{-5mm}
\end{figure}

\paragraph{Setup} We randomly sample 1000 topics learned by topic models, then use different prompts to refine the topics with LLAMA3-8B-Instruct. We analyze the effectiveness of prompts in different aspects, including \textbf{Success Rate} \citep{ulmer-etal-2024-calibrating}: the proportion of cases where the target answer can be successfully extracted from the LLM's output; \textbf{N\_Input} and \textbf{N\_Output} \citep{chang2024enhanced}: the average number of tokens of input and output of the LLM; and \textbf{Refined TC}: the average topic coherence scores of the refined topics.

\paragraph{Results} From the results in Table \ref{tab:prompt_performance}, we observe the following: (1) Through prompt optimization, the effectiveness of the prompt can be further enhanced (e.g., Variant\_4), where the number of tokens (i.e., the cost) is reduced and the refine topics are more coherent. (2) The iterative refinement \citep{chang2024enhanced} shows less effectiveness in terms of both cost and refined topic coherence compared with our prompt variants when applied to LLAMA3-8B-Instruct. 

Based on the above observations, we further investigate the effectiveness of the improved prompt within the LLM-ITL framework. We plot the learning curves of LLM-ITL using the original prompt and its variant (Variant\_4, which shows better performance from Table \ref{tab:prompt_performance}). We observe that the overall performance in terms of both metrics is comparable by using both prompts.

\begin{table*}[t]
  \centering
  \resizebox{\textwidth}{!}{
  \begin{tabular}{p{1.5cm}|p{20cm}}
    \toprule
     & \textbf{Prompt} \\
    \midrule
    Origin & Analyze step-by-step and provide the final answer.

Step 1. Given a set of words, summarize a topic (avoid using proper nouns as topics) by 2 words that covers most of those words. Note, only the topic, no other explanations.

Step 2. Remove irrelevant words about the topic from the given word list. Note, only the removed words, no other explanations.

Step 3. Add new relevant words (maximum 10 words) about the topic to the word list up to 10 words. Note, only the added words, no other explanations.

Step 4. Provide your answer in json format as \{``Topic'': ``\textless2 Word Topic\textgreater'', ``Words'': ``\textless Refined 10 Word List\textgreater''\}. Note, only 10 refined words allowed for the topic, and no follow up explanations.\\
\cmidrule(l){1-2}
Variant\_1 & Perform the following actions sequentially and provide the final result:

Step 1. After examining a set of words, condense a subject (avoid proper nouns) into 2 words that encompass most of those words. (Note: Only the subject, no further elaboration.)

Step 2. Eliminate irrelevant words from the given word list based on the subject. (Note: Only the removed words, no further elaboration.)

Step 3. Add new pertinent words (maximum 10 words) related to the subject to the word list until it reaches 10 words. (Note: Only the added words, no further elaboration.)

Step 4. Present your response in JSON format as \{``Topic'': ``\textless2 Word Subject\textgreater'', ``Words'': ``\textless Refined 10 Word List\textgreater''\}. Note: Only 10 refined words are permitted for the subject, and no follow-up explanations. \\
\cmidrule(l){1-2}
Variant\_2 & Perform a meticulous examination and furnish the conclusive resolution.

Stride 1. Bestowed a catalogue of vocabularies, condense a subject matter (circumvent the employment of proper appellations as subjects) by dual words that envelop the preponderance of those vocabularies. (Heed, solely the subject, devoid of supplemental explication.)

Stride 2. Dislodge irrelevant vocabularies concerning the subject from the granted vocabulary catalogue. (Heed, solely the dislodged vocabularies, devoid of supplemental explication.)

Stride 3. Amalgamate novel applicable vocabularies (maximal 10 vocabularies) concerning the subject to the vocabulary catalogue up to 10 vocabularies. (Heed, solely the amalgamated vocabularies, devoid of supplemental explication.)

Stride 4. Tender your resolution in json format as \{``Topic'': ``\textless 2 Word Subject\textgreater'', ``Words'': ``\textless Refined 10 Word Catalogue\textgreater''\}. Heed, solely 10 refined vocabularies permitted for the subject, and devoid of successive explication.\\
\cmidrule(l){1-2}
Variant\_3 & Step-by-step analysis and final answer:

Step 1. Given a set of words, summarize a topic (avoid using proper nouns as topics) by 2 words that covers most of those words. (Note, only the topic, no other explanations.)

Step 2. Remove irrelevant words about the topic from the given word list. (Note, only the removed words, no other explanations.)

Step 3. Add new relevant words (maximum 10 words) about the topic to the word list, keeping the total word count at 10 words. (Note, only the added words, no other explanations.)

Step 4. Provide your answer in JSON format as \{``Topic'': ``\textless2 Word Topic\textgreater'', ``Words'': ``\textless Refined 10 Word List\textgreater''\}. Note, only 10 refined words allowed for the topic, and no follow-up explanations.\\
\cmidrule(l){1-2}
Variant\_4 & Break down the analysis into steps and give the final response.

1. Look at a set of words and identify a 2-word topic that sums up most of those words (don't use proper nouns as topics, just state the topic).

2. Remove words from the list that don't relate to the topic (just list the removed words).

3. Add new relevant words about the topic to the list, up to 10 words total (just list the new added words).

4. Provide your response in JSON format: \{``Topic'': ``\textless 2 Word Topic\textgreater'', ``Words'': ``\textless Refined 10 Word List\textgreater''\}. Only include 10 words for the refined list, no explanations.\\
\cmidrule(l){1-2}
Variant\_5 & Step-by-step analysis and provide the final answer in JSON format:

Step 1: Based on the given set of words, summarize a topic using 2 words that encompass most of those words (avoid proper nouns).

Step 2: Remove any irrelevant words from the given word list that do not relate to the summarized topic.

Step 3: Add new relevant words (up to 10 words) that are related to the summarized topic.

Step 4: Present your answer in the following JSON format: \{``Topic'': ``\textless 2 Word Topic\textgreater'', ``Words'': ``\textless Refined 10 Word List\textgreater''\}, where ``Topic'' contains the 2-word summarized topic, and ``Words'' contains the refined list of 10 words related to that topic. Do not provide any additional explanations.\\
\cmidrule(l){1-2}
Iterative Refinement \citep{chang2024enhanced} & Please analyze the following tasks and provide your answer in the specified format.

1. Determine the common topic shared by these words:
[\textless TOPIC\_WORDS \textgreater ].

2. Assess whether the word ``\textless WORD\textgreater'' aligns with the same
common topic as the words listed above.

Respond with:

- ``Yes'', if the given word shares the common topic.

- If ``No'', suggest 10 single-word alternatives that are commonly
used and closely related to this topic. These words should be easily
recognizable and distinct from the ones in the provided list.

Format your response in JSON, including the fields ``Topic'',
``Answer'', and ``Alternative words'' (only if the answer is ``No'').\\
  \bottomrule
\end{tabular}
}
  \caption{Prompt variants for topic refinement}
  \label{prompt_variants}
\end{table*}

\section{Topic Quality Based on Distributed Word Representations}\label{sec:extra_coherence}
\setcounter{table}{0} 
\renewcommand{\thetable}{I\arabic{table}}
\begin{table}[H]
    \centering
    \resizebox{0.5\textwidth}{!}{
    \begin{tabular}{clrrr}
    \toprule
    \textbf{Dataset} & \textbf{Method} & \textbf{W2V-Cosine} $(\downarrow)$ & \textbf{W2V-L1} $(\downarrow)$ & \textbf{W2V-L2} $(\downarrow)$\\
    \midrule
      \multirow{2}{*}{20News} & ETM & 0.218\small{ ± 0.002} & 2.313\small{ ± 0.017} & 11.459\small{ ± 0.180}\\
      & + LLM-ITL & 0.177\small{ ± 0.005} & 2.128\small{ ± 0.040} & 9.650\small{ ± 0.335} \\
      & & \bm{$\uparrow$} \textbf{18.8\%} & \bm{$\uparrow$} \textbf{8.0\%} & \bm{$\uparrow$} \textbf{15.8\%}\\
    \cmidrule(l){1-5}
    \multirow{2}{*}{R8} & ETM & 0.237\small{ ± 0.004} & 2.543\small{ ± 0.023} & 13.673\small{ ± 0.211} \\
    & + LLM-ITL & 0.180\small{ ± 0.006} & 2.245\small{ ± 0.042} & 10.855\small{ ± 0.345}\\
    & & \bm{$\uparrow$} \textbf{24.1\%} & \bm{$\uparrow$} \textbf{11.7\%} & \bm{$\uparrow$} \textbf{20.6\%}\\
    \cmidrule(l){1-5}
    \multirow{2}{*}{DBpedia} & ETM & 0.213\small{ ± 0.003} & 2.448\small{ ± 0.020} & 12.709\small{ ± 0.237}\\
    & + LLM-ITL & 0.163\small{ ± 0.010} & 2.119\small{ ± 0.090} & 9.633\small{ ± 0.696}\\
    & & \bm{$\uparrow$} \textbf{23.5\%} & \bm{$\uparrow$} \textbf{13.4\%} & \bm{$\uparrow$} \textbf{24.2\%}\\
\cmidrule(l){1-5}
    \multirow{2}{*}{AGNews} & ETM & 0.203\small{ ± 0.005} & 2.419\small{ ± 0.023} & 12.382\small{ ± 0.234}\\
    & + LLM-ITL & 0.164\small{ ± 0.004} & 2.175\small{ ± 0.035} & 10.031\small{ ± 0.241}\\
    & & \bm{$\uparrow$} \textbf{19.2\%} & \bm{$\uparrow$} \textbf{10.1\%} & \bm{$\uparrow$} \textbf{19.0\%}\\
       \bottomrule
    \end{tabular}
    }
\caption{Topic quality in terms of Word2vec metrics.}
\label{tab:extra_coherence}
\end{table}

In this section, we conduct a performance comparison of topic quality using metrics based on distributed word representations \citep{10.1145/2911451.2914720}, which have been shown to align better with human judgment than standard topic coherence metrics.

Specifically, we use the topic quality metrics based on word vectors as implemented by \citet{10.1145/2911451.2914720}, applying cosine distance (\textbf{W2V-Cosine}), L1 distance (\textbf{W2V-L1}), and L2 distance (\textbf{W2V-L2}), respectively, as the distance functions in the calculation. As shown in Table \ref{tab:extra_coherence}, LLM-ITL consistently improves semantic coherence of topics, as measured by the Word2Vec-based evaluation metrics.

\end{document}